\documentclass[10pt,journal,compsoc]{IEEEtran}
\usepackage{array}
\usepackage[caption=false,font=normalsize,labelfont=sf,textfont=sf]{subfig}
\usepackage{stfloats}
\usepackage{url}
\usepackage{verbatim}
\usepackage{cite}
\usepackage{amsmath,amssymb,amsfonts}
\usepackage{algorithmic}
\usepackage{graphicx}
\usepackage{textcomp}
\usepackage{xcolor}
\usepackage{multirow}
\usepackage{booktabs}
\usepackage{algorithm}
\newcommand{\eg}{\textit{e.g.}}
\newcommand{\ie}{\textit{i.e.}}
\newtheorem{definition}{Definition}
\newtheorem{theorem}{Theorem}[section]
\newtheorem{lemma}[theorem]{Lemma}
\newtheorem{proposition}[theorem]{Proposition}

\DeclareMathOperator*{\argmin}{arg\,min}
\DeclareMathOperator*{\argmax}{arg\,max}
\usepackage{dsfont}

\usepackage{hyperref}
\usepackage{tablefootnote}

\def\BibTeX{{\rm B\kern-.05em{\sc i\kern-.025em b}\kern-.08em
    T\kern-.1667em\lower.7ex\hbox{E}\kern-.125emX}}
\begin{document}


\title{MARIO: Model Agnostic Recipe for Improving OOD Generalization of Graph Contrastive Learning}

\author{
Yun~Zhu,
Haizhou~Shi,
Zhenshuo~Zhang,
Siliang~Tang,~\IEEEmembership{Member,~IEEE},

\markboth{Journal of \LaTeX\ Class Files,~Vol.~14, No.~8, August~2021}%
{Shell \MakeLowercase{\textit{et al.}}: A Sample Article Using IEEEtran.cls for IEEE Journals}




\IEEEcompsocitemizethanks{
    \IEEEcompsocthanksitem Yun Zhu, Zhenshuo Zhang and Siliang Tang are with the Department of Computer Science, Zhejiang University, Hangzhou, China.\protect\\
    Email: \{zhuyun$\_$dcd, zs.zhang, siliang\}@zju.edu.cn
    \IEEEcompsocthanksitem Haizhou Shi is in the Department of Computer Science, Rutgers University, New Brunswick, US.\protect\\
    Email: haizhou.shi@rutgers.edu
}%
}

\IEEEtitleabstractindextext{
\begin{abstract}
In this work, we investigate the problem of out-of-distribution (OOD) generalization for unsupervised learning methods on graph data. This scenario is particularly challenging because graph neural networks (GNNs) have been shown to be sensitive to distributional shifts, even when labels are available.
To address this challenge, we propose a \underline{M}odel-\underline{A}gnostic \underline{R}ecipe for \underline{I}mproving \underline{O}OD generalizability of unsupervised graph contrastive learning methods, which we refer to as MARIO. MARIO introduces two principles aimed at developing distributional-shift-robust graph contrastive methods to overcome the limitations of existing frameworks: (i) Information Bottleneck (IB) principle for achieving generalizable representations and (ii) Invariant principle that incorporates adversarial data augmentation to obtain invariant representations.
To the best of our knowledge, this is the first work that investigates the OOD generalization problem of graph contrastive learning, with a specific focus on node-level tasks. Through extensive experiments, we demonstrate that our method achieves state-of-the-art performance on the OOD test set, while maintaining comparable performance on the in-distribution test set when compared to existing approaches. The source code for our method can be found at: \hyperlink{https://github.com/ZhuYun97/MARIO}{https://github.com/ZhuYun97/MARIO}.
\end{abstract}

}

\maketitle
\IEEEdisplaynontitleabstractindextext
\IEEEpeerreviewmaketitle

\section{Introduction}
\IEEEPARstart{G}{raph-structured} data is pervasive in various aspects of our lives, such as social networks~\cite{community}, citation networks~\cite{citation}, and molecules~\cite{molecule}. Most existing graph learning algorithms~\cite{gcn,sage,gin,gat,8294302,8519335} work under the statistical assumption~\cite{erm,ermxxx} that the training and testing data are drawn from the same distribution. However, in real-world scenarios, this assumption does not often hold. For example, in citation networks, as time progresses, new topics emerge and citation distributions change~\cite{ogb,eerm,good}, which consequently causes a distribution shift between the training and testing domains. Apart from this challenge of training-test distributional shift, how to effectively utilize the massive unannotated graph data emerging every day also remains to be an intriguing problem to solve. Hence, the primary goal of this paper is to seek for a set of principles that achieves superior OOD generalization performance with unlabeled graph data. To this end, two primary challenges need to be addressed:

\emph{Challenge 1:} Non-Euclidean data structure of graphs causes complex distributional shifts (feature-level and topology-level) and lack of environment labels (due to the inherent abstraction of graph), which in consequence severely qualifies the direct application of existing OOD generalization methods.

\emph{Challenge 2:} Most existing OOD generalization methods heavily rely on label information. It remains a practical challenge how to elicit invariant representations when no access to labels is provided.

Many efforts have been made towards the resolution of the challenges above. To address \emph{Challenge~1}, EERM~\cite{eerm}, GIL~\cite{gil}, and DIR~\cite{dir} employ environment generators to simulate diverse distributional shifts in graph data. By minimizing the mean and variance of risks across multiple graphs and environments, these methods manage to capture invariant features that generalize well on unseen domains. However, these approaches heavily rely on the label information, which cannot be deployed in unsupervised settings, as \emph{Challenge~2} suggests. Regarding the second challenge, graph contrastive learning (GCL) has recently emerged as a prominent unsupervised graph learning framework. Although some of the GCL methods have demonstrated superior performance under in-distribution tests, their efficacy under out-of-distribution tests is still unclear, as they do not explicitly target on improving the OOD generalization ability. In summary, current methods struggle to effectively address both challenges simultaneously in the field of unsupervised OOD generalization for graph data.

In this work, we for the first time systematically study the robustness of current unsupervised graph learning methods~\cite{rosa,grace,graphcl,bgrl,dgi,mvgrl,gmae,costa} while facing distribution shifts.
By analyzing the common drawbacks of GCL methods, we propose a \textbf{M}odel-\textbf{A}gnostic \textbf{R}ecipe for \textbf{I}mproving \textbf{O}OD generalization of GCL methods (MARIO\footnote{Based on this recipe, we provide a shift-robust graph contrastive framework coined as MARIO}). To solve the above challenges, MARIO works on the two crucial components of a typical GCL method, \ie, view generation and representation contrasting, as depicted in Figure~\ref{fig:contrast}, and leaves the encoding models as an open choice for existing and future works for the sake of universal application. 
Concretely, MARIO introduces two principles aimed at developing distributional-shift-robust graph contrastive methods to overcome the limitations of existing frameworks: (i) Information Bottleneck (IB) principle for achieving generalizable representations (solving \emph{Challenge 2}) and (ii) Invariance principle that incorporates adversarial data augmentation to acquire invariant representations (solving \emph{Challenge 1}).
Throughout extensive experiments, we observe that some graph contrastive methods are more robust to distribution shift, especially in datasets with artificial spurious features (\eg, GOOD-CBAS). Furthermore, our proposed model-agnostic recipe MARIO reaches comparable performance on the in-distribution test domain but shows superior performance on out-of-distribution test domain, regardless of what model is deployed for view encoding.

As the first work that investigates the efficacy of unsupervised graph learning methods while facing distribution shifts~\footnote{In this work, we specifically focus on node-level downstream tasks, which are more challenging than graph-level tasks due to the interconnected nature of instances within a graph.}, our paper's main contributions are  summarized as follows:
\begin{itemize}
    \item Through extensive experiments, we observe that some GCL methods are more robust to OOD tests than their supervised counterparts, providing insights for solving the challenge of graph OOD generalization.
    \item Motivated by invariant learning and information bottleneck, we analyze the limitations of the main components in current GCL frameworks for OOD generalization, and we further propose a \textbf{M}odel-\textbf{A}gnostic \textbf{R}ecipe for \textbf{I}mproving \textbf{O}OD generalization of GCL methods (MARIO). 
    \item The proposed model-agnostic recipe MARIO can be seamlessly deployed for various graph encoding models, achieving SOTA performance under the out-of-distribution test while reaching comparable performance under the in-distribution test.
\end{itemize}

\section{Background and Problem Formulation~\label{sec:pre}}
In this section, we will start with the notations we use throughout the rest of the paper~(Sec.~\ref{sec:notations}); then we introduce the problem definition and background of graph OOD generalization~(Sec.~\ref{sec:graph-ood}) and graph self-supervised learning methods~(Sec.~\ref{sec:ssl-graph}); finally, we formalize the problem of graph contrastive learning for OOD generalization~(GCL-OOD) in Sec.~\ref{sec:gcl-ood}. 

\subsection{Notations}\label{sec:notations}
\noindent Let $\mathcal{G},\mathcal{Y}$ represent input and label space respectively. $f_\phi(\cdot)=p_\omega\circ g_\theta(\cdot)$ represents graph predictor which consists of a GNN encoder $g_\theta(\cdot)$ and a classifier $p_\omega(\cdot)$. The graph predictor $f_\phi: \mathcal{G}\rightarrow\mathcal{Y}$ maps instance $G=(A,X) \in \mathcal{G}$ to label $Y \in \mathcal{Y}$ where $A \in \mathbb{R}^{N \times N}$ is the adjacent matrix and $X \in \mathbb{R}^{N \times D}$ is the node attribute matrix. Here, $N$, $D$ denote the number of nodes and attributes, respectively. To measure the discrepancy between the prediction and the ground-truth label, a loss function $\ell_{\text{sup}}$ is used (\eg, cross-entropy loss, mean square error). For unsupervised learning, a pretext loss $\ell_{\text{unsup}}$ is applied (\eg, InfoNCE loss~\cite{cpc}). And we use $\mathcal{T}$ as augmentation pool, the augmentation function $\tau$ is randomly selected from $\mathcal{T}$ according to some distribution $\pi$. Let $\mathcal{G}^{\text{tar}}$ denote downstream dataset. 

\subsection{Graph OOD Generalization}\label{sec:graph-ood}
\textbf{Problem definition.} Given training set $\mathcal{G}^{\text{train}}={(G_i,Y_i)}_{i=1}^N$ containing $N$ instances that are drawn from training distribution $P_{\text{train}}(G,Y)$. In the supervised setting, it aims to learn an optimal graph predictor $f^*$ that can exhibit the best generalization performance on the data sampled from the test distribution:
\begin{equation}
f_\phi^*=\arg \min _{f_\phi} \mathbb{E}_{G, Y \sim P_{\text {test }}}\left[\ell_{\text{sup}}\left(f_\phi(G), Y\right)\right],
\end{equation}
where $P_{\text {test }}(G, Y) \neq P_{\text {train }}(G, Y)$ means there exists a distribution shift between training and testing sets. Such a shift may lead to the optimal predictor trained on the training set (\ie, minimizing $\mathbb{E}_{G, Y \sim P_{\text {train }}}\left[\ell_{\text{sup}}\left(f_\phi(G), Y\right)\right]$) can not generalize on the testing set.

\noindent\textbf{Related works.} 
Out-of-distribution generalization (aka domain generalization) algorithms~\cite{9782500,8496795,ood-survey,good-survey}, developed to handle unknown distribution shifts, have gained significant attention due to the growing need for handling unseen data in real-world scenarios\footnote{There are similar topics like domain adaptation~\cite{zhu2021shift,wu2020unsupervised} and transfer learning~\cite{zhu2021transfer,5288526}. They typically assume access to part of test domains to adapt GNNs. In contrast, domain generalization does not use any samples from test domains, setting it apart from these approaches.}. Specifically, robust optimization~\cite{groupdro,hu2018does}, invariant representation/predictor learning~\cite{irm,sparseirm} and causal approaches~\cite{peters2016causal,heinze2018causal} are proposed to deal with such problems. In this subsection, we will put an emphasis on \emph{invariant representation learning} on graphs because of its more practical assumption and theory.

Graph invariant learning methods extend invariant learning on graph domain which are widely investigated recently~\cite{eerm,gil,dir,gsat}. EERM \cite{eerm}, GIL\cite{gil}, DIR~\cite{dir} rely on environment generators to find invariant predictive patterns with labels. GSAT~\cite{gsat} leverages the attention mechanism and the information bottleneck principle~\cite{dib} to construct interpretable GNNs for learning invariant subgraphs under distribution shifts. But most works focus on graph-level tasks under supervised setting. 

Dealing with node-level tasks without labels is more challenging due to the interconnected graph structure and the large scale of individual graphs as well as the lack of supervision. Training a model with good generalization in this scenario remains under-explored. In this work, we aim to develop an OOD algorithm for node-level tasks without labels, addressing this challenging problem.

\subsection{Graph Contrastive Learning}\label{sec:ssl-graph}
\begin{figure}
    \centering
    \includegraphics[scale=0.5]{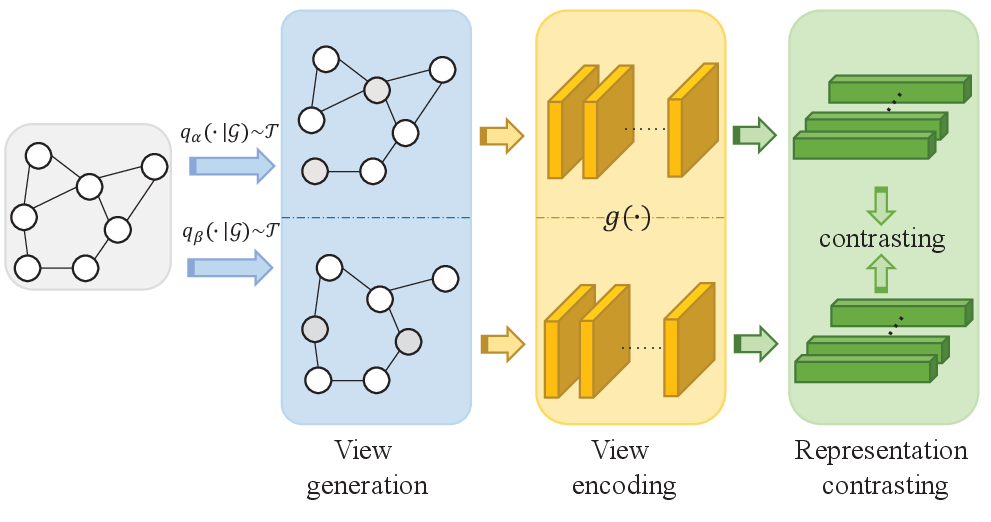}
    \caption{The pipeline of graph contrastive learning.}
    \label{fig:contrast}
\end{figure}

\textbf{Problem definition.} 
Graph contrastive learning (GCL) is a representative self-supervised graph learning method~\cite{rosa,grace,graphcl,costa,bgrl,dgi,mvgrl,9764632}. It consists of three main components: view generation, view encoding, and representation contrasting (Figure~\ref{fig:contrast}). Given an input graph $G$, two graph augmentations $\tau_{\alpha}$ and $\tau_{\beta}$ are used to generate two augmented views $G_\alpha=\tau_\alpha(G)$ and $G_\beta=\tau_\beta(G)$, respectively. A GNN model $g_\theta$~\cite{gcn,gat,gin,sage} is then applied to the augmented views to produce node representations $g_\theta(\tau_{\alpha}(G))\in\mathbb{R}^{N \times D}$.
Lastly, a contrastive loss function is applied to the representations, pulling together the positive pairs  while pushing apart negative pairs. Taking the commonly used InfoNCE loss~\cite{cpc} as an example, the formulation follows:
\begin{equation}
\begin{aligned}
\mathcal{L}_{\text{MI}} & \left(g_\theta;\mathcal{G},\pi\right)=-\underset{G \in \mathcal{G}}{\mathbb{E}} \mathbb{E} _{\tau_\alpha, \tau_\beta \sim \pi^2}\left\|g_\theta(\tau_\alpha(G))-g_\theta\left(\tau_\beta(G)\right)\right\|^2 \\
+ & \underset{G \in \mathcal{G}}{\mathbb{E}} \log \underset{G^{\prime} \in \mathcal{G}}{\mathbb{E}}\mathbb{E}_{\tau^{\prime} \sim \pi}\left[  e^{\left\|g_\theta\left(\tau_\alpha\left(G\right)\right)-g_\theta\left(\tau^{\prime}(G^\prime)\right)\right\|^2}\right],
\end{aligned}
\label{equ:con}
\end{equation}
where in the second term, $G^\prime$ denotes a randomly sampled graph from the graph data distribution $\mathcal{G}$, serving as the constraint for non-collapsing representations.
For simplicity, the representation produced by encoder is automatically normalized to a unit sphere, \ie,  $\|g_\theta(G)\|=1, \forall G \in \mathcal{G}$. By minimizing this loss, the former term (aka alignment loss $\mathcal{L}_{\text{align}}$~\cite{wang2020understanding}) pulls positive pairs together by encouraging their similarity, and the latter term (aka uniformity loss $\mathcal{L}_{\text{uniform}}$~\cite{wang2020understanding}) pushes negative pairs apart. 
The quality of the pre-trained graph encoder is then evaluated by the linear separability of the final representations. Namely, an additional trainable linear classifier $p_\omega$ is built on top of the frozen encoder:
\begin{equation}
p_\omega^*=\arg \min _{p_\omega} \mathbb{E}_{G, Y \sim P_{\text {train }}}\left[\ell_{\text{sup}}\left(p_\omega \circ g_\theta^*(G), Y\right)\right],
\end{equation}
where $g_\theta^*(\cdot)$ is obtained by minimizing Equation~\ref{equ:con} without labels. For evaluating pre-trained model, the optimal graph predictor $f_\phi^* =  p_\omega^* \circ g_\theta^*$ will be applied to testing data. 

\noindent\textbf{Related work.} Recently, contrastive methods in the graph domain have shown remarkable progress, surpassing supervised methods even without human annotations in certain cases~\cite{gmae,dgi,graphcl,grace,rosa,mvgrl,bgrl,zhu2023sgl,zhang2023structure,raft,9770382}. These self-supervised methods mostly assume an in-distribution (ID) setting, where the training and test data are sampled from the identical distribution. However, in real-world scenarios, where there are distribution shifts between training and test sets, the efficacy of existing methods remains a question. To solve this, RGCL~\cite{rgcl} employs a rationale generator to find casual subgraph for each instance~(graph) as graph augmentation in contrastive learning for further improving the OOD generalization performance. However, RGCL cannot be applied to node-level tasks because the rationale generator cannot be applied to large-scale graphs due to memory consumption and each instance~(node) is intractable to find its rational subgraph.

In this work, we are the first to investigate the robustness of graph self-supervised methods in the face of distribution shifts on node-level tasks. We provide a model-agnostic recipe for improving the OOD generalization of graph contrastive learning methods which does not rely on the choice of models. By addressing this research question, we provide valuable insights and practical guidelines for improving the performance of unsupervised graph contrastive learning with distribution shifts. In the next chapter, we will explicitly formulate the problem of graph contrastive learning for OOD generalization~(GCL-OOD) and pinpoint the corresponding challenges.

\subsection{GCL-OOD: Graph Contrastive Learning for OOD Generalization}\label{sec:gcl-ood}
Suppose $\Phi(G)$ is invariant rationales of input instance $G$ which is stable in different environments (augmentations) following invariance assumption~\cite{ood-survey,good-survey}: 
\begin{equation}
    \mathbb{E}\left[Y \mid \Phi\left(G_e\right)\right]=\mathbb{E}\left[Y \mid \Phi\left(G_{e^{\prime}}\right)\right], \quad \forall e,e^\prime \in \operatorname{supp}\left(\mathcal{E}_{t r}\right),
\end{equation}
where $\mathcal{E}_{t r}$ denotes the set of training environments\footnote{Data in different environments has different data distributions.} and the above equation represents that invariant rationales exhibit predictive invariant (stable) correlations with semantic labels across different environments.

The optimal~(invariant) graph encoder $g_\theta^\star$ achieves the invariant rationales $\Phi(G)$ across all the environments: \footnote{We assume the augmentation function will not change the semantic labels of the original input here.}:
\begin{equation}
    g_\theta^{\star}(G_e)=g_\theta^{\star}(G_{e^\prime}) = \Phi(G),  \quad \forall e,e^\prime \in \operatorname{supp}\left(\mathcal{E}_{t r}\right).
\label{equ:goal}
\end{equation}
However, during the pre-training of GCL, we have no access to labels under self-supervised setting. Here, we build a connection between pre-text loss $\mathcal{L}_{\text{MI}}\left(g_\theta ; \mathcal{G}, \pi\right)$ and downstream loss $\mathcal{R}\left(g_\theta ; \mathcal{G}^{\operatorname{tar}}\right)$ by upper-bounding referring to~\cite{arcl,huang2023towards}:
\begin{equation}
\begin{aligned}
\mathcal{R}\left(p_\omega \circ g_\theta ; \mathcal{G}_\pi\right) \leq 
& c\|p_\omega\| \sqrt{K \sigma}\left(\mathcal{L}_{\text {align }}(g_\theta ; \mathcal{G}, \pi)\right)^{\frac{1}{4}} \\ 
& +\|p_\omega\| \zeta (\sigma, \delta) \\
& +\sum_{k=1}^K \mathcal{G}_\pi\left(C_k\right)\left\|e_k-p_\omega \circ \mu_k\left(g_\theta ; \mathcal{G}_\pi\right)\right\|,
\end{aligned}
\label{equ:cl_down}
\end{equation}
where $c$ is a positive constant, $\zeta (\sigma, \delta)$ is a set of constants that only depends on $(\sigma, \delta)$-augmentation~\cite{huang2023towards}, $C_k \subseteq \mathcal{G}$ is the set of the data points in class $k$, $\mu_k(g_\theta ; \mathcal{G}):=\mathbb{E}_{G \sim \mathcal{G}}[ g_\theta(G)] \text { for } k \in[K]$. The complete derivation and more illustrations can be found in Appendix~\ref{app:proof_recipe1}. 

The first term in Equation~\ref{equ:cl_down} is the alignment loss optimized during pre-training on $\mathcal{G}$. 
The second term is determined by the $(\sigma,\delta)$ quantity of the data augmentation, with larger $\sigma$ and smaller $\delta$ resulting in smaller $\zeta(\sigma,\delta)$.
The third term is associated with the linear layer $p$ and is minimized in downstream training. The class centers can be distinguished by choosing an appropriate regularization term $\mathcal{L}_{\text{uniform}}$, leading to the third term becoming 0 via $h$.
In short, Equation~\ref{equ:cl_down} implies that contrastive learning on distribution $\mathcal{G}$ with augmentation function $\tau$ essentially optimizes the upper-bound of supervised risk on the augmented distribution $\mathcal{G}_\tau$ resulting in a lower supervised risk. So, even without labels, we can approach the goal formulated as Equation~\ref{equ:goal} during pre-training to some extent, through modifying the main components in current GCL methods which will be discussed in Section~\ref{sec:method}.

For evaluating the pre-trained models, the linear evaluation protocol mentioned in the last subsection will be used: 
\begin{equation}
p_\omega^*=\arg \min _{p_\omega} \mathbb{E}_{G, Y \sim P_{\text {train }}}\left[\ell_{\text{sup}}\left(p_\omega \circ g_\theta^\star(G), Y\right)\right],
\end{equation}
where pre-trained model $g_\theta^\star$ is frozen, and the linear classifier $p_\omega$ is trained with training data\footnote{We train the linear head using the training data which is more appropriate to OOD generalization setting, rather than using partial test data as in \cite{shi2022robust}.}. The difference with the last subsection is $P_{\text {test }}(G, Y) \neq P_{\text {train }}(G, Y)$ which means there exist distribution shifts between training and testing data. A good performance on the test data distribution implies that the feature extractor must have extracted invariant features, mitigating the risk of overfitting on the training data and leading to improved generalization under distribution shifts.

\section{Shift-Robust Graph Contrastive Learning~\label{sec:method}}
\noindent In this section, we will provide a \textbf{M}odel-\textbf{A}gnostic \textbf{R}ecipe for \textbf{I}mproving \textbf{O}OD generalizability of GCL methods, dubbed MARIO. 
A GCL training pipeline can be typically decomposed into three components: (i)~view generation, (ii)~view encoding, and (iii)~representation contrasting, as illustrated in Figure~\ref{fig:contrast}. 
MARIO works on the first~(view generation) and the last component~(representation contrasting), leaving the view encoding as an orthogonal design choice for GCL methods. Therefore it can be agnostically applied to different graph encoding models such as GCN~\cite{gcn}, GAT~\cite{gat}, GraphSAGE~\cite{sage} and etc. 

In the remaining content, we will first analyze the drawbacks of the two components in existing GCL methods for OOD generalization and introduce our proposed recipe correspondingly in Section~\ref{sec:aug} and Section~\ref{sec:cmi}. Finally, we will formulate the complete training scheme for graph OOD generalization problem in Section~\ref{sec:training}. \textbf{The complete derivation and more detailed illustration of all lemmas, theorems and corollaries can be found in the Appendix.}

\subsection{Recipe 1: Revisiting Graph Augmentation~\label{sec:aug}}
\noindent Data augmentation plays a crucial role in the transferability and generalization ability of contrastive learning~\cite{simclr,tian2020makes,huang2023towards,arcl}. It is proved that contrastive learning on distribution $\mathcal{G}$ with augmentation function $\tau$ essentially optimizes the supervised risk on the augmented distribution $\mathcal{G}_\tau$ instead of the original distribution $\mathcal{G}$~\cite{arcl}.
Consequently, if the downstream distribution $\mathcal{G}^{\mathrm{tar}}$ is similar to training distribution $\mathcal{G}$, the encoder obtained by contrastive learning shall perform well on it. 
Although the alignment loss in the contrastive learning achieves certain level of generalization, the learned representation distribution lacks domain invariance since it only takes expectation over the augmentation distribution $\pi$~\cite{arcl}. 
This limitation will severely hinder the OOD generalization ability of models~\cite{arcl,irm}. 

Our first improvement of graph contrastive learning based on graph augmentation is motivated by invariant learning~\cite{irm,arcl} which aims to learn domain-invariant features across $\{\mathcal{G}_{\tau}\}_{\tau \in \mathcal{T}}$ to solve \emph{Challenge 1}. Firstly, let us retrospect invariant risk minimization~\cite{irm}.

\begin{definition}[Invariant risk minimization,~IRM] If there is a classifier $p_{\omega^*}$ simultaneously optimal for all domains in $\mathcal{B}$, we will say that a data representation $g_\theta$ elicits an invariant predictor $p_{\omega^*} \circ g_\theta$ across a domain set $\mathcal{B}$:
\begin{equation}
p_{\omega^*} \in \arg \min _{p_\omega} \mathcal{R}(p_\omega \circ g_\theta ; \mathcal{G}) \text { for all } \mathcal{G} \in \mathcal{B},
\end{equation}
\label{def:irm}
where $\mathcal{R}$ denotes the risk of the predictor $p_\omega \circ g_\theta$ measured on domain $\mathcal{G}$.
\end{definition}

Definition~\ref{def:irm} yields the features that exhibit stable correlations with the target variable. It has been empirically and theoretically demonstrated that such features can enhance the generalization of models across distribution shifts in supervised learning~\cite{irm,ahuja2021invariance,li2022invariant}.
By setting $\mathcal{B}$ to the set of augmented graphs $\{\mathcal{G}_\tau\}_{\tau \in \mathcal{T}}$, this concept can be readily applied to graph contrastive learning methods with~\cite{groupdro,arcl}. The following definition of invariant alignment loss is the proposed objective for GCL-OOD problem, and we will draw the connection between Definition~\ref{def:irm} and Definition~\ref{def:invariant-alignment} in Theorem~\ref{thm:upper-bound}.

\begin{definition}[Invariant Alignment Loss]\label{def:invariant-alignment}
The invariant alignment loss $\mathcal{L}_{\mathrm{align}^*}$ of the graph encoder $g_\theta$ over the graph distribution $\mathcal{G}$ is defined as
\begin{equation}
\mathcal{L}_{\mathrm{align}^*}(g_{\theta} ; \mathcal{G}):=\underset{G \in \mathcal{G}}{\mathbb{E}} \sup _{\tau, \tau^{\prime} \in \mathcal{T}}\left\|g_{\theta}(\tau(G))-g_{\theta}\left(\tau^{\prime}(G)\right)\right\|^2.
\end{equation}
\end{definition}

The invariant alignment loss measures the difference between two representations under the most ``challenging'' two augmentations, rather than the trivial expectation as in Equation~\ref{equ:con}. Intuitively, it avoids the situation where the encoder behaves extremely differently in different $\mathcal{G}_\tau$. A special case of binary classification problem is analysed in Appendix \ref{app:case} to substantiate it.
Then we will discuss why the supremum operator can solve such a dilemma.
\begin{theorem}[Upper bound of variation across different domains~\cite{arcl}]\label{thm:upper-bound}
    For two augmentation functions $\tau$ and $\tau^{\prime}$, linear predictor $p$ and representation $g$, the variation across different domains is upper-bounded by
    \begin{equation}
        \sup _{\tau, \tau^{\prime} \in \mathcal{T}}\left|\mathcal{R}\left(p \circ g ; \mathcal{G}_\tau\right)-\mathcal{R}\left(p \circ g ; \mathcal{G}_{\tau^{\prime}}\right)\right| \leq c \cdot\|p\| \mathcal{L}_{\mathrm{align}^*}(f, \mathcal{G}) .
    \end{equation}
    Furthermore, fix $g$ and let $p_\tau \in \arg \min _p \mathcal{R}\left(p \circ g, \mathcal{G}_\tau\right)$. Then we have
    \begin{equation}
    \begin{aligned}
        |\mathcal{R}\left(p_\tau \circ g ; \mathcal{G}_{\tau^{\prime}}\right) & - \mathcal{R}\left(p_{\tau^{\prime}} \circ g ; \mathcal{G}_{\tau^{\prime}}\right)| \leq \\ & 2 c \cdot\left(\left\|p_\tau\right\|+\left\|p_{\tau^{\prime}}\right\|\right) \mathcal{L}_{\mathrm{align^*}}(g, \mathcal{G}).
    \end{aligned}
    \label{equ:proposition}
    \end{equation}
    \label{theorem:align}
\end{theorem}
The complete deduction and the connection between contrastive loss and downstream risk $\mathcal{R}$ can be found in Appendix~\ref{app:proof_recipe1}.
$\mathcal{L}_{\text{align}^*}$ replace the expectation operator over $\mathcal{T}$ with the supremum operator in $\mathcal{L}_{\text{align}}$ of Equation~\ref{equ:con}
resulting in $\mathcal{L}_{\text {align }}(g ; \mathcal{G}, \pi) \leq \mathcal{L}_{\text{align}^*}(g ; \mathcal{G})$ for all $g$ and $\pi$, and the augmentation function $\tau$ is randomly selected from the augmentation pool $\mathcal{T}$ according to a certain distribution $\pi$. 
When $\mathcal{L}_{\text {align}^*}$ is optimized to a small value, it indicates that $\mathcal{R}\left(p \circ g ; \mathcal{G}_\tau\right)$ varies a little under different augmentation functions $\tau$ resulting in the optimal representation for $\mathcal{G}_\tau$ is close to $\mathcal{G}_{\tau^\prime}$. That is, representation with smaller $\mathcal{L}_{\text {align}^*}$ tends to elicit the same linear optimal predictors across different domains, a property that does not hold for original alignment loss $\mathcal{L}_{\text {align}}$. In addition to pulling positive pairs together, $\mathcal{L}_{\text {align}^*}$ can ensure that this alignment is consistent and uniform across $\{\mathcal{G_\tau}\}_{\tau \in \mathcal{T}}$~\cite{arcl}.
 
\textbf{Adversarial augmentation.}
One problem with replacing $\mathcal{L}_{\text {align}}$ with  $\mathcal{L}_{\text {align}^*}$ is that it is intractable to estimate $\sup _{\tau, \tau^{\prime} \in \mathcal{T}}\left\|g(\tau(G))-g\left(\tau^{\prime}(G)\right)\right\|^2$, since it requires us to iterate over all augmentation methods. Previous work\cite{arcl} adopts multiple views and selects the worst pair to approximate the supermum operator. However, this strategy is straightforward and the quality is heavily dependent on the number of views. In order to find the worst case in the continuous space efficiently, we turn to the adversarial training~\cite{shafahi2019adversarial,flag,kim2020adversarial,suresh2021adversarial} to approximate the supermum operator:
\begin{equation}
\min _{\theta} \mathbb{E}_{(G, Y) \sim \mathcal{G}}\left[\max _{\|{\delta}\|_p \leq \epsilon} L\left(g_{\theta}(X+{\delta}, A), Y\right)\right],
\end{equation}
where the inner loop maximizes the loss to approximate the most challenging perturbation, whose strength $\|\delta\|\leq \epsilon$ is strictly controlled so that it does not change the semantic labels of the original view, \eg, $\epsilon=1e-3$.
Considering the training efficiency, in this paper, we follow and further modify the supervised graph adversarial training framework FLAG \cite{flag} to accommodate unsupervised graph contrastive learning as follows:
\begin{equation}
\min _{\theta} \mathbb{E}_{(G_\alpha, G_\beta) \sim \mathcal{G}}\left[\max _{\|{\delta}\|_p \leq \epsilon} L\left(g_{\theta}(X_\alpha+{\delta}, A_\alpha), g_{\theta}(X_\beta, A_\beta)\right)\right].
\label{equ:unsup_ad}
\end{equation}

Note that it is not new for graph learning models to employ adversarial augmentation~\cite{kim2020adversarial,jiang2020robust,rosa}, our proposed recipe significantly differs from the previous works in terms of the objective: they focus on enhancing the robustness of models against adversarial attacks, while we leverage adversarial augmentation to specifically improve the OOD generalization of contrastive learning methods. By providing theoretical justifications and further elucidating the underlying principles, we contribute to a deeper understanding of the benefits of adversarial augmentation for OOD generalization in the contrastive learning paradigm.

\subsection{Recipe 2: Revisiting Representations Contrasting~\label{sec:cmi}} 
\begin{figure}
    \centering
    \includegraphics[scale=0.5]{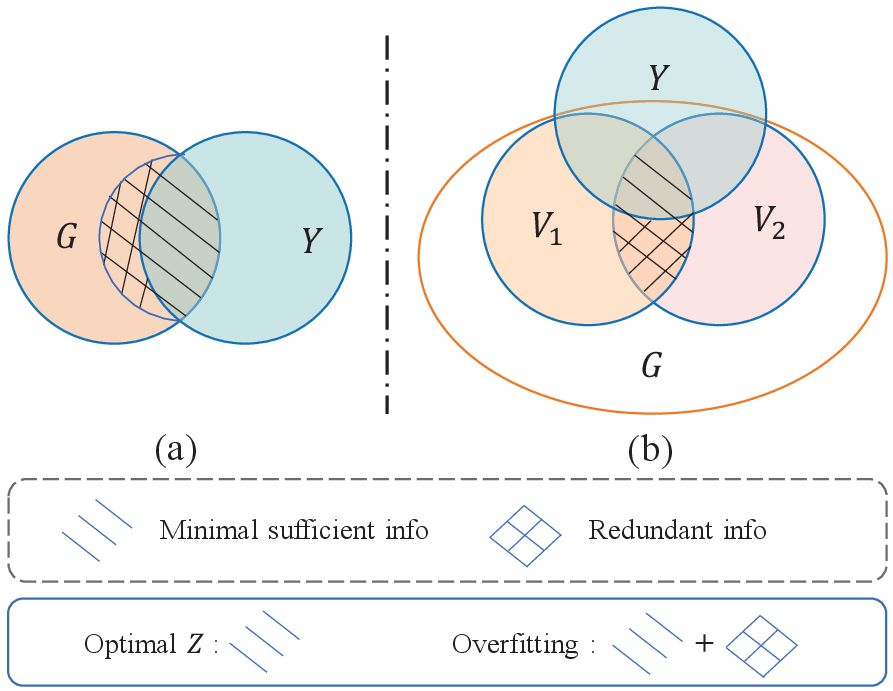}
    \caption{Venn diagram of mutual information and conditional mutual information}
    \label{fig:mi}
\end{figure}

\noindent The vanilla contrastive loss like Equation~\ref{equ:con} aims to maximize the lower bound of the mutual information between positive pairs. However, there exists some redundant information (\ie, conditional mutual information) that can impede the generalization of graph contrastive learning. 
Our objective is to learn minimal sufficient representation related to downstream task which can effectively mitigate overfitting and demonstrate robustness against distribution shifts.
In this subsection, we introduce a recipe for representation contrasting to improve the generalization of GCL methods, motivated by the principle of information bottleneck~\cite{tishby_ib} to solve \emph{Challenge 2}. In short, we refer to the modified contrastive loss to get rid of supervision signals as well as learning generalized representations.
\begin{definition}[Information Bottleneck, IB]
    Let $X, Z, Y$ represent random variables of inputs, embeddings, and labels respectively. The formulation of information bottleneck's training objective is 
    \begin{equation}
    \arg\max_{\theta} R_{I B}(\theta)=I_{{\theta}}(Z; Y)-\beta I_{{\theta}}(Z; X),
    \label{equ:ib2}
    \end{equation}
    where $I_\theta$ represents mutual information estimator with parameters $\theta$, and $\beta > 0$ controls the trade-off between compression and the downstream task performance (larger $\beta$ leads to lower compression rate but high MI between the embedding $Z$ and the label $Y$). 
    \end{definition}
The definition above implies that the IB principle~\cite{dib,tishby_ib} aims to learn the minimal sufficient representation for the given task by maximizing the mutual information between the representation and the target (\emph{sufficiency}), and simultaneously constraining the mutual information between the representation and the input data (\emph{minimality}) as depicted in Figure~\ref{fig:mi}a.
By adopting this learning paradigm, the trained model can effectively mitigate overfitting and exhibit improved resilience against distribution shifts~\cite{gib,li2022invariant,ahuja2021invariance}. 

Motivated by this principle, we modify the vanilla contrastive loss~\cite{grace,rosa,costa} as Equation~\ref{equ:cmi}. 
The current graph contrastive learning methods aim to maximize the mutual information between positive pairs as depicted in Figure~\ref{fig:mi}b. Nevertheless, in scenarios where training labels are accessible, some shared information in the vanilla contrastive loss is redundant. In Figure~\ref{fig:mi}b, $V_1, V_2$ represents two augmented views from the same sample $\mathcal{G}$ and $U, V$ represent the representations of $V_1, V_2$, the redundant information which should be eliminated based on the IB principle. 
To precisely describe the redundant information here, let us introduce the concept of conditional mutual information~(CMI).


\begin{definition}[Conditional Mutual Information, CMI] 
The conditional mutual information $I(U; V\mid Y)$ measures the expected value of mutual information between $U$ and $V$ given $Y$ which can be formulated as 
\label{def:cmi}
\begin{equation}
\begin{aligned}
\operatorname{I}(U ; V \mid Y):&=\mathbb{E}_{y \sim Y}\left[D_{\mathrm{KL}}\left(P_{U, V \mid Y=y} \| P_{U \mid Y=y} P_{V \mid Y=y}\right)\right]\\&=\int_{\mathcal{Y}} D_{\mathrm{KL}}\left(P_{U, V \mid Y} \| P_{U \mid Y} P_{V \mid Y}\right) \mathrm{d} P_Y.
\end{aligned}
\end{equation}
\end{definition}
To reduce the redundant information and hence improve the OOD generalization ability, we need to minimize the CMI between two views $U$ and $V$. However, it is intractable to estimate the equation above. In this work, we appeal to mutual information estimators~(\eg, Donsker-Varadhan estimator~\cite{donsker1975asymptotic,mine}, Jensen-Shannon estimator~\cite{fgan,mine}, InfoNCE~\cite{nce,cpc}) to estimate the lower bound of conditional mutual information.
Taking InfoNCE~\cite{cpc} as an example, the CMI objective can be approximated as
\begin{equation}
\begin{aligned}
\mathcal{L} _ {CMI} & \left(U, V\right)=-\mathbb{E}_{y \sim P_Y}\Big[\mathbb{E}_{u,v \sim P_{U,V\mid y}}\left[\operatorname{sim}\left(u, v\right)\right] \\
+ & \mathbb{E}_{u \sim P_{U \mid y}} \log \mathbb{E}_{v^- \sim P_{V\mid y}} \left[  e^{\operatorname{sim}\left(u, v^-\right)}\right]\Big]
\end{aligned}
\label{equ:con_cmi}
\end{equation}
where $\operatorname{sim}(x,y)$ is the cosine similarity function, the positive pairs are drawn from the conditional joint distribution, and negative pairs are drawn from the product of conditional marginal distribution. In short, we first sample $y \sim Y$, and then we sample positive and negative pairs from $P_{U,V \mid y}$ and $P_{U \mid y} P_{V \mid y}$\footnote{A classifier is required to determine which category the representations of the augmented inputs belong to. The true labels can be used for training the classifier.}.
The negative format of Equation~\ref{equ:con} is a lower bound of the conditional mutual information $I(U;V \mid Y)$. The proof can be found in Appendix~\ref{app:cmi}.

\textbf{Online Clustering.} The main challenge of applying the above approximation to our unsupervised pre-training lies in lack of labels~$Y$. To address this issue, we utilize online clustering techniques to obtain pseudo labels. These pseudo labels are iteratively refined during training, ensuring an increase in their mutual information with the ground-truth labels~\cite{pcl}. In order to incorporate the clustering into our pre-text task, we employ similar strategies as in \cite{swav,pcl}. We will initialize learnable prototypes $\boldsymbol{c}_i$ for each cluster $i$ and matrix $C=\left[c_1,c_2,\cdots,c_K\right]$ collects all column prototype vectors. For clustering, we can simply calculate the similarity between $K$ prototypes and node representations $u_i$ and $v_i$ for node~$i$:
\begin{equation}
\begin{aligned}
    &p_{u_i}\left(\hat{y} \mid u_i\right)=\operatorname{softmax}\left(C^T \cdot u_i\right),\\
    &q_{v_i}\left(\hat{y} \mid v_i\right)=\operatorname{softmax}\left(C^T \cdot v_i\right),
    \label{equ:pq}
\end{aligned}
\end{equation}
where the prototypes $C$ are updated by solving the problem of swapped prediction~\cite{swav}:
\begin{equation}
\begin{aligned}
    &\mathcal{L}_{\mathrm{clu}}\left(U,V\right)=\sum_i^B\left[\ell\left(p_{u_i},q_{v_i}\right)+\ell\left(q_{u_i}, p_{v_i}\right)\right], \\
    &\text{where} \quad \ell\left(p_{u_i}, q_{v_i}\right)=-\sum_k q_{v_i}^{(k)} \log p_{u_i}^{(k)}.
\end{aligned}
\label{equ:clu}
\end{equation}
The clustering loss focuses on contrasting nodes from different views by comparing cluster assignments rather than their representations.
However, there exists a trivial solution that all data samples are allocated to the same cluster. This problem can be solved by introducing another constraint of equal partition of the prototype assignment~\cite{swav}. Please refer to Appendix~\ref{app:online} for details.

For stable training, we use bi-level optimization~\cite{bilevel} for updating the encoder and prototypes (more details in Section~\ref{sec:training}). 
With these prototypes, we can infer the pseudo labels of node representations:
\begin{equation}
    \hat{Y} = \argmax C^TU.
\end{equation}
Hence our final shift-robust contrastive loss can be formulated as
\begin{equation}
\begin{aligned}
    \min_{g_\theta}\mathcal{L}_{\mathrm{rob}}&=\argmax I_{\theta}(U;V) - \gamma I_{\theta}(U;V \mid \hat{Y}) \\
                              &=\argmin_{g_\theta} \mathcal{L}_{\text{MI}} - \gamma \mathcal{L}_{\text{CMI}},
\end{aligned}
\label{equ:cmi}
\end{equation}
where $I_{\boldsymbol{\theta}}(U;V), I_{\boldsymbol{\theta}}(U;V \mid \hat{Y})$ can be instantiated as Equation~\ref{equ:con} and Equation~\ref{equ:cmi} respectively; $\gamma \ge 0$ controls the trade-off between compression and pre-text task's performance similar in Equation~\ref{equ:ib2}.
Intuitively, if the positive pairs have already shared the same semantic labels in the feature space (\ie, belong to the same cluster), the objective will reduce their shared information to avoid learning redundant information and overfitting~\cite{gib,dib} during training, which will bring performance gain in OOD generalization.

\subsection{Model Training~\label{sec:training}}\noindent Our shift-robust contrastive loss is formulated as Equation~\ref{equ:cmi} which involves maximizing the mutual information and minimizing the conditional mutual information at the same time. As mentioned in Section~\ref{sec:cmi}, bi-level optimization~\cite{bilevel} is used for updating the prototypes $C$ and other parameters in Equation~\ref{equ:cmi}:

\begin{equation}
\begin{aligned}
g^{k+1} &=\arg\min_g \mathcal{L}_{\text{rob}}(g^k, C^{k+1}; \mathcal{G}, \mathcal{T}), \\
C^{k+1} &=\arg\min_C \mathcal{L}_{\text{clu}}(g^k,C; \mathcal{G}, \mathcal{T}),
\end{aligned}
\end{equation}
where the parameter of the graph encoder $g_\theta$ is omitted for uncluttered notations. 
Concretely, we first fix the encoder $g$ and update the prototypes with $l$ steps of SGD to approximate the $\arg\min$ optimization, and then with the near-optimal prototypes $C$, we update the parameters of the encoder for $1$ step of SGD.
With the adversarial augmentation, our final optimization objective $\mathcal{L}_{\text{rob}}$ for the graph encoder is replaced by:
\begin{gather}
 \min _{g} \mathbb{E}_{(G_\alpha, G_\beta) \sim \mathcal{G}}\left[\max _{\|{\delta}\|_p \leq \epsilon} \mathcal{L}_{\text{rob}}\left(g(X_\alpha+{\delta}, A_\alpha), g(X_\beta, A_\beta), C\right)\right].
\end{gather}
The algorithm can be summarized as in Algorithm~\ref{alg1}. The combination of the Invariant principle (Recipe 1) and the Information Bottleneck principle (Recipe 2) results in learned representations with enhanced OOD generalization, as demonstrated in the supervised setting~\cite{li2022invariant}. The ablation study in Section~\ref{sec:ablation} further confirms the effectiveness of this combination, leading to a synergistic effect where the whole is greater than the sum of its parts.
\begin{algorithm}[t]
    \textbf{Input}: Augmentation pool $\mathcal{T}$, ascent steps $M$, ascent step size $\epsilon$, encoder $g_\theta$, projector $p_\omega$, and training graph $G=(A, X)$ \\
    \begin{algorithmic}[1] 
        \WHILE{not converge}
        \STATE $\tau_{\alpha}, \tau_{\beta} \sim \mathcal{T}$
        \STATE $G_\alpha, G_\beta = \tau_{\alpha}(G),\tau_{\beta}(G)$
        \STATE $C=\arg\min_C\mathcal{L}_{\text{clu}} \left(g(G_\alpha), g(G_\beta), C\right)$
        
        \STATE $\delta_{0} \leftarrow U(-\epsilon, \epsilon)$
        \STATE $\hbar_{0} \leftarrow 0$
        \FOR{$\mathrm{t}=1 \ldots M$} 
        \STATE $Z_\alpha=p_{{\omega}} \circ g_{\theta}\left(X_\alpha+\delta_{t-1} , A_\alpha\right)$
        \STATE $Z_\beta = p_\omega \circ g_\theta(X_\beta, A_\beta)$
        \STATE $\hbar_{t} \leftarrow \hbar_{t-1}+\frac{1}{M} \cdot \nabla_{\theta, \omega} \mathcal{L}_{\text{rob}}\left(Z_\alpha, Z_\beta, C\right)$
        \STATE $\hbar_{\delta} \leftarrow \nabla_{\delta} \mathcal{L}_{\text{rob}}\left(Z_\alpha, Z_\beta, C\right)$
        \STATE $\delta_{t} \leftarrow \delta_{t-1}+\epsilon \cdot \hbar_{\delta} /\left\|\hbar_{\delta}\right\|_{F}$
        \ENDFOR
        \STATE $\theta \leftarrow \theta-\eta \cdot \hbar_{M, \theta}$
        \STATE $\omega \leftarrow \omega-\eta \cdot \hbar_{M, \omega}$
        \ENDWHILE
    \end{algorithmic}
    \caption{Algorithm for a shift-robust framework for graph contrastive learning.}
    \label{alg1}
\end{algorithm}

It is important to note that our optimization approach remains efficient. The optimization process of prototypes involves only a few steps (\eg, 10 steps in our experiments) of gradient descent, and the number of parameters involved is relatively small. Regarding adversarial augmentation, the inner loop iterates a moderate number of times, such as 3 iterations in our experiments, to approximate the optimal perturbation. During this process, we accumulate gradients in the inner loop to update the parameters in the outer loop. The time complexity of our method is linearly proportional to that of GRACE~\cite{grace}. Therefore, the additional computational overhead introduced by our optimization approach is acceptable.
\section{Experiments}
In this section, we first introduce the experimental setup including datasets, training, and evaluation protocol in Section~\ref{sec:dataset}~and~\ref{sec:unsupervised}. 
We then perform an ablation study to demonstrate the effectiveness of each proposed component in Section~\ref{sec:ablation}. 
Additionally, we analyze the impact of important hyper-parameters in Section~\ref{sec:sensitivity}. 
Subsequently, we integrate our method with various encoding models, showcasing the model-agnostic nature of our recipe in Section~\ref{sec:other_models}. 
Finally, we provide some qualitative results such as feature visualization in Section~\ref{sec:vis}.
It is important to note that we focus on node-level tasks (\eg, node classification) in this work. As for graph-level tasks, we leave it as our future work, while some simple experiments are also provided in Appendix~\ref{app:graph_classification}.

\subsection{Datasets}\label{sec:dataset}
There exist some benchmarks for evaluating graph out-of-distribution generalization~\cite{good,ji2022drugood,gds}. 
Among them, GOOD~\cite{good} is the most representative and comprehensive benchmark that curates more diverse graph datasets with diverse tasks, including single/multi-task graph classification, graph regression, and node classification involving more distribution shifts (\ie, concept shifts and covariate shifts). Hence in this work, we follow the evaluation protocol proposed in \cite{good}. Furthermore, we validate the effectiveness of our method in the datasets (\ie, Amazon-Photo, Elliptic) that are used in EERM~\cite{eerm}. The statistics and detailed introduction to these datasets can be found in Table~\ref{tab:dataset} and Appendix~\ref{app:datasets}.

\begin{table*}[htp]
\caption{The descriptions of datasets. ``Domain-Level'' means splitting by graphs, ``Time-Aware'' denotes splitting according to chronological order.``Word'' and ``Degree'' represent splitting according to word diversity and node degree respectively. ``Language'' means splitting by user language, suggesting the prediction should not be impacted by the language the user use. ``University'' denotes splitting according to the domain university, implying that the prediction of webpages should be based on word contents and link connections rather than university features. ``Color'' means that nodes are split according to node differences in covariate shift and color-label correlations in concept shift.}
\label{tab:dataset}
\centering
\begin{tabular}{cccccccc}
\toprule
Datasets     & Network Type        & \#Nodes & \#Edges & \#Attributes &\#Classes& Train/Val/Test Split     & Metric   \\
Amazon-Photo\footnotemark
             & Co-purchasing network      & 7,650   & 119,081   & 755          & 10      & Domain-Level         & Accuracy \\
Elliptic\footnotemark  
             & Bitcoin transactions       & 203,769 & 234,355   & 165          & 2       & Time-Aware           & F1-Score \\
GOOD-Cora    & Scientific publications    & 19,793  & 126,842   & 8,710         & 70      & Word/Degree          & Accuracy \\
GOOD-Twitch  & Gamer network              & 34,120  & 892,346   & 128          & 2       & Language             & ROC-AUC  \\
GOOD-CBAS    & A BA-house graph           & 700     & 3,962     & 4             & 4       & Color                & Accuracy \\
GOOD-WebKB   & Webpage network            & 617     & 1,138     & 1,703         & 5       & University           & Accuracy \\
\bottomrule
\end{tabular}
\end{table*}
\footnotetext[5]{This dataset is adopted from~\cite{yang2016revisiting}. \cite{eerm} constructs ten graphs with different environment id’s for each graph.} 
\footnotetext[6]{The original is available on \hyperlink{https://www.kaggle.com/ellipticco/elliptic-data-set}{https://www.kaggle.com/ellipticco/elliptic-data-set}}

\subsection{Unsupervised Representation Learning}\label{sec:unsupervised}
\subsubsection{Transductive Setting}\label{sec:trans}
In this subsection, we focus on validating our proposed algorithm under the transductive setting, where the test nodes will participate in message passing~\cite{gilmer2017neural} during training following~\cite{good}. 

\noindent\textbf{Baselines.} We conduct experiments with 12 baselines from three categories: (i)~supervised methods, including empirical risk minimization~(\textbf{ERM})~\cite{erm}, invariant risk minimization (\textbf{IRM})~\cite{irm}, and a recent proposed graph OOD method \textbf{EERM}~\cite{eerm}; (ii)~self-supervised generative methods including Graph Autoencoder (\textbf{GAE})~\cite{gae}, Variational Graph Autoencoder (\textbf{VGAE})~\cite{gae}, Self-Supervised Masked Graph Autoencoders (\textbf{GraphMAE})~\cite{gmae}; (iii)~self-supervised contrastive methods including Deep Graph Infomax (\textbf{DGI})~\cite{dgi}, Contrastive Multi-View Representation Learning on Graphs (\textbf{MVGRL})~\cite{mvgrl}, Deep Graph Contrastive Representation Learning (\textbf{GRACE})~\cite{grace}, A Robust Self-Aligned Framework for Node-Node Graph Contrastive Learning (\textbf{RoSA})~\cite{rosa}, Bootstrapped Representation Learning on Graphs (\textbf{BGRL})~\cite{bgrl}, Covariance-Preserving Feature Augmentation for Graph Contrastive Learning (\textbf{COSTA})~\cite{costa}, Unsupervised Learning of Visual Features by Contrasting Cluster Assignments (\textbf{SwAV})~\cite{swav}. The detailed descriptions of these baselines can be found in Appendix~\ref{app:baselines}.

\noindent\textbf{Experimental setup.} We use the same graph encoder across different datasets for a fair comparison following~\cite{good}. We use grid search to find other hyper-parameters (\eg, learning rate, epochs) for different methods. For all experiments, we select the best checkpoints for ID and OOD tests according to results on ID and OOD validation sets following~\cite{good}, respectively. Experimental details and hyper-parameter selections are provided in Appendix~\ref{app:hyper}. For evaluating unsupervised methods, a linear classifier will be built on the frozen trained encoder after finishing pre-training. The reported results are the mean performance with standard deviation after 10 runs following~\cite{good}.

\noindent\textbf{Analysis.}\quad Based on the experimental results listed in Table~\ref{tab:trans_concept} and \ref{tab:trans_covariate}, we can draw the following conclusions: firstly, we find strong self-supervised methods (\eg, GRACE, BGRL, COSTA) are more robust to distribution shifts (concept shift in Table~\ref{tab:trans_concept} and covariate shift in Table~\ref{tab:trans_covariate}) compared to supervised methods. For instance, on GOOD-CBAS and GOOD-WebKB datasets, GRACE surpasses the best supervised method by large margins (over 6\% absolute improvement). Interestingly, we find the methods designed for OOD generalization (\ie, IRM) and graph OOD generalization (\ie, EERM) do not attain superior performance than the standard ERM on most of the datasets. For example, EERM shows superior OOD performance compared to ERM in only one experiment, and IRM outperforms ERM in four out of ten experiments across the conducted evaluations. This phenomenon is also observed in \cite{good,ahuja2020empirical,rosenfeld2021risks}, showcasing the challenge of achieving invariant prediction in non-Euclidean graph settings. 

Furthermore, our method surpasses other SOTA self-supervised methods on the OOD test set of all datasets by a considerable margin while achieving comparable performance in the in-distribution test set. For instance, on small datasets such as GOOD-CBAS and GOOD-WebKB, our method outperforms GRACE\footnote{MARIO is built up on GRACE according to our recipe. So, we make a comparison with GRACE here.} by over 2\% absolute accuracy on the OOD test set. On larger datasets such as GOOD-Cora and GOOD-Twitch, our method still outperforms other methods which shows its superiority. For instance, under covariate shift, MARIO surpasses other methods by over 7\% absolute accuracy on the GOOD-Twitch OOD test set. These statistics prove the effectiveness of our design.

\begin{table*}[htp]
\caption{Experimental results of all methods under concept shift. The bold font means the top-1 performance and the underline represents the second performance across the unsupervised methods. 'ID' represents in-distribution test performance and 'OOD' means out-of-distribution test performance. (OOM: out-of-memory on a GPU with 24GB memory)}
\label{tab:trans_concept}
\centering
\scalebox{0.95}{
\begin{tabular}{l|cc|cc|cc|cc|cc}
\toprule
\toprule
\multirow{3}{*}{concept shift} & \multicolumn{4}{c|}{GOOD-Cora}                   & \multicolumn{2}{c|}{GOOD-CBAS} & \multicolumn{2}{c|}{GOOD-Twitch} & \multicolumn{2}{c}{GOOD-WebKB} \\
                           & \multicolumn{2}{c}{word} & \multicolumn{2}{c|}{degree}& \multicolumn{2}{c|}{color}    & \multicolumn{2}{c|}{language}   & \multicolumn{2}{c}{university} \\
                           & ID         & OOD         & ID          & OOD          & ID            & OOD           & ID             & OOD            & ID            & OOD            \\
\midrule
ERM                        & 66.38±0.45 & 64.44±0.18  & 68.60±0.40  & 60.76±0.34   & 89.79±1.39    & 83.43±1.19    & 80.80±1.00     & 56.92±0.92     & 62.67±1.53    & 26.33±1.09     \\
IRM                        & 66.42±0.41 & 64.29±0.31  & 68.57±0.35  & 61.45±0.24   & 89.64±1.21    & 82.29±1.14    & 78.87±1.04     & 59.30±1.79     & 62.67±1.10    & 26.88±1.42     \\
EERM                       & 65.10±0.44 & 62.45±0.19  & 66.95±0.44  & 56.58±0.25   & 79.07±2.12    & 64.50±1.01    & OOM            & OOM            & 62.50±2.01    & 28.07±3.23      \\
\midrule
GAE                        & 60.65±0.89 & 58.00±0.55  & 62.59±1.11  & 53.44±0.80   & 75.28±1.36    & 68.07±2.05    & 81.25±0.81     & 51.51±1.05     & 62.17±3.34    & 25.78±1.85     \\
VGAE                       & 63.19±0.53 & 60.35±0.47  & 61.65±0.66  & 54.28±0.28   & 76.50±0.50    & 59.07±0.56    & 80.46±0.53     & 55.56±4.53     & 62.50±2.38    & 24.40±2.57     \\
GraphMAE                   & \underline{66.44±0.46} & \underline{64.87±0.30}  & 67.95±0.46  & 59.41±0.39   & 89.14±0.89    & 82.93±0.93    & 80.05±0.64     & 59.38±1.49     & 61.83±3.37    & 29.27±2.15     \\
DGI                        & 63.33±0.56 & 60.71±0.49  & 65.93±1.02  & 55.83±0.53   & 91.22±1.47    & 85.00±1.66    & 80.05±0.87     & 59.16±1.88     & 61.83±2.83    & 28.63±1.92      \\
MVGRL                      & OOM        & OOM         & OOM         & OOM          & 88.57±1.15    & 76.50±1.17    & OOM            & OOM            & 62.00±3.79    & 28.26±4.20     \\
GRACE                      & 65.61±0.61 & 63.92±0.44  & \textbf{68.59±0.35}  & 60.15±0.45   & 92.00±1.39    & 88.64±0.67    & \textbf{83.43±0.63}     & \underline{60.45±1.46}     & 64.00±3.43    & \underline{34.86±3.43}  \\
RoSA                       & 64.06±0.67 & 62.44±0.39  & 67.07±0.65  & 57.68±0.44   & 90.78±2.27    & 85.93±2.14    & 82.39±0.42     & 57.45±2.16     & 64.17±4.10    & 32.20±2.15     \\
BGRL                       & 65.18±0.43 & 63.43±0.45  & 66.83±0.80  & 59.63±0.38   & 92.36±1.16    & 87.14±1.60    & 82.52±0.60     & 55.48±1.48     & 63.67±2.33    & 31.47±3.43     \\
COSTA                      & 65.05±0.80 & 62.37±0.45  & 66.76±0.87  & 55.73±0.36   & \underline{93.50±2.62}    & \underline{89.29±3.11}    & 83.15±0.30 & 55.03±3.22     & 61.66±2.58    & 32.39±2.13 \\
SwAV                       & 62.22±0.53 & 59.79±0.53  & 64.65±0.94  & 55.06±0.39   & 89.00±0.79    & 81.72±0.66    & \underline{83.32±0.15}     & 59.69±1.97     & \underline{65.17±3.76}    & 29.36±2.01    \\
\midrule
MARIO                       & \textbf{67.11±0.46} & \textbf{65.28±0.34}  & \underline{68.46±0.40}  & \textbf{61.30±0.28}   & \textbf{94.36±1.21}    & \textbf{91.28±1.10}    & 82.31±0.54     & \textbf{63.33±1.72}     & \textbf{65.67±2.81}    & \textbf{37.15±2.37}     \\
\bottomrule
\end{tabular}}
\end{table*}

\begin{table*}[htp]
\caption{Experimental results of all methods under covariate shift. The bold font means the top-1 performance and the underline represents the second performance across the unsupervised methods. 'ID' represents in-distribution test performance and 'OOD' means out-of-distribution test performance. (OOM: out-of-memory on a GPU with 24GB memory)}
\label{tab:trans_covariate}
\centering
\scalebox{0.95}{
\begin{tabular}{l|cc|cc|cc|cc|cc}
\toprule
\toprule
\multirow{3}{*}{covariate shift} & \multicolumn{4}{c|}{GOOD-Cora}                                   & \multicolumn{2}{c|}{GOOD-CBAS} & \multicolumn{2}{c|}{GOOD-Twitch} & \multicolumn{2}{c}{GOOD-WebKB} \\
                           & \multicolumn{2}{c}{word} & \multicolumn{2}{c|}{degree}& \multicolumn{2}{c|}{color}    & \multicolumn{2}{c|}{language}   & \multicolumn{2}{c}{university} \\
                           & ID         & OOD         & ID          & OOD          & ID            & OOD           & ID             & OOD            & ID            & OOD            \\
\midrule
ERM                        & 70.50±0.41 & 64.69±0.33  & 72.46±0.49  & 55.53±0.50   & 92.00±3.08    & 77.57±1.29    & 70.98±0.41     & 49.35±5.09     & 39.34±1.79    & 14.52±3.14   \\
IRM                        & 70.48±0.26 & 64.53±0.57  & 71.98±0.34  & 53.72±0.46   & 90.86±2.41    & 78.86±1.67    & 69.81±0.95     & 49.11±2.82     & 38.52±3.30    & 13.97±2.80     \\
EERM                       & OOM        & OOM         & OOM         & OOM          & 65.00±2.57    & 57.43±3.60    & OOM            & OOM            & 46.07±4.55    & 27.40±7.65     \\
\midrule
GAE                        & 56.63±0.79 & 48.93±0.93  & 66.30±0.88  & 34.01±0.87   & 73.00±2.16    & 60.86±3.01    & 67.24±1.23     & 47.65±2.49     & 45.08±6.32    & 28.02±6.29    \\
VGAE                       & 62.02±0.66 & 54.12±0.86  & 69.41±0.57  & 44.20±1.29   & 62.29±2.04    & 63.29±1.11    & 66.99±1.43     & \underline{50.48±4.58}     & 48.85±4.68    & 20.87±6.69     \\
GraphMAE                   & 68.14±0.43 & 64.00±0.33  & \textbf{73.36±0.56}  & 53.75±0.55   & 67.28±3.03    & 67.28±1.49    & 68.84±1.20     & 48.02±2.79     & 48.03±4.34    & 30.00±8.09     \\
DGI                        & 60.85±0.75 & 57.03±0.67  & 68.97±0.41  & 41.75±0.88   & 69.57±4.09    & 59.71±3.43    & 68.43±1.05     & 44.83±1.61     & 48.52±5.04    & 21.11±7.50     \\
MVGRL                      & OOM        & OOM         & OOM         & OOM          & 65.00±1.94    & 64.15±0.77    & OOM            & OOM           & \textbf{54.10±5.39}    & 16.59±6.51     \\
GRACE                      & \underline{68.77±0.33} & \underline{64.21±0.41}  & 72.69±0.34  & \underline{56.10±0.63}   & \underline{93.57±1.83}    & \underline{89.29±3.40}    & \underline{71.12±0.87} & 46.21±1.54 & 49.67±5.82    & 28.10±4.68    \\
RoSA                       & 68.19±0.56 & 62.48±0.61  & 71.04±0.62  & 52.72±0.79   & 84.71±4.14    &79.14±3.51     & 70.58±0.36     & 45.83±1.72     & 52.30±4.24    & \underline{34.24±7.92}     \\
BGRL                       & 67.23±0.43 & 61.33±0.36  & 72.11±0.39  & 49.15±0.73   & 89.00±2.56    & 79.86±3.29    & \textbf{71.43±0.53}     & 43.86±0.94     & 51.80±5.55    & 30.32±7.61    \\
COSTA                      & 65.28±0.60 & 60.33±0.53  & 70.65±0.62  & 54.03±0.28   & 92.29±1.59    & 82.71±2.74    & 69.29±1.37     & 49.07±2.13     & 50.49±3.01    & 29.84±4.75   \\
SwAV                       & 63.29±1.01 & 56.98±0.94  & 70.27±0.73  & 43.00±0.52   & 89.57±1.12    & 81.43±1.69    & 69.19±0.93     & 49.37±2.96     & 49.84±4.82    & 30.55±6.72   \\
\midrule
MARIO                       & \textbf{69.99±0.54} & \textbf{65.06±0.34}  & \underline{72.73±0.43}  & \textbf{57.73±0.45}  & \textbf{94.57±2.46}    & \textbf{91.00±2.48}     & 68.31±0.78 & \textbf{57.37±1.37}     & \underline{53.94±3.23}    & \textbf{35.24±4.98}   \\
\bottomrule
\end{tabular}}

\end{table*}

\subsubsection{Inductive Setting}
In this subsection, we conduct experiments under the inductive settings, where the test nodes are kept unseen during training. This setting is more suitable for domain generalization.

\noindent\textbf{Baselines:} For GOOD-WebKB and GOOD-CBAS datasets, we adopt ERM, IRM, GraphMAE, and GRACE as our baselines. And for Amazon-Photo and Elliptic datasets, we select ERM, EERM, and GRACE as our baselines.

\noindent\textbf{Experimental setup:} For GOOD-WebKB and GOOD-CBAS datasets, we use the same model configuration in Section~\ref{sec:trans}.
For Amazon-Photo dataset~\cite{yang2016revisiting} and Elliptic~\cite{elliptic} dataset, they consist of many snapshots (training data and testing data use different snapshots) which are naturally inductive. For Amazon-Photo dataset, we use 2-layer GCN~\cite{gcn} as the encoder and for elliptic dataset, we use 5-layer GraphSAGE~\cite{sage} as encoder following~\cite{eerm}.

\begin{figure*}[htp]
    \centering
    \includegraphics[scale=0.45]{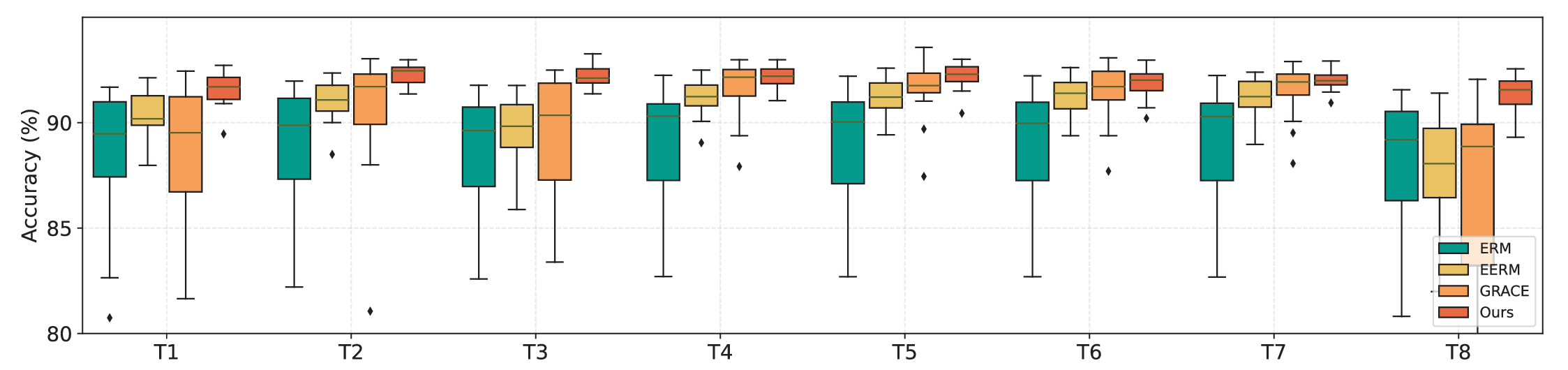}
    \caption{Results on Amazon-photo dataset with artificial distribution shifts. T1$\sim$T8 represents different test graphs which are created with different environment IDs.}
    \label{fig:amazon}
\end{figure*}

\begin{figure}[htp]
    \centering
    \includegraphics[scale=0.4]{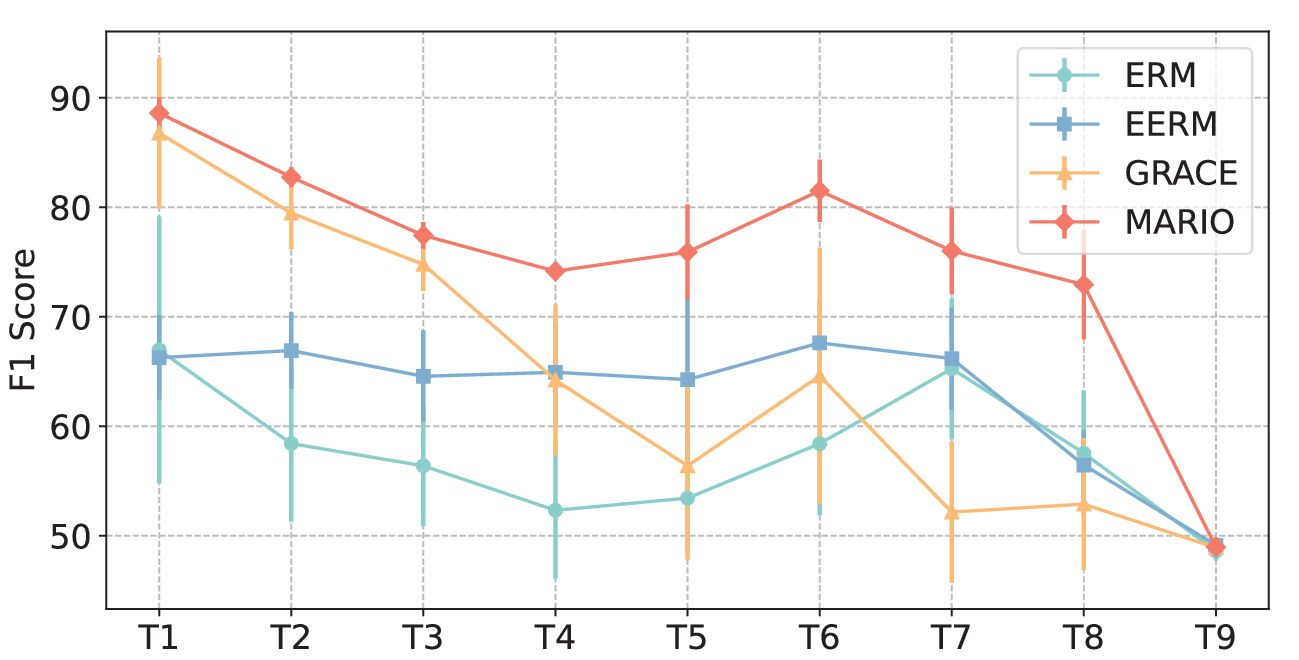}
    \caption{Experiment results on Elliptic dataset with label distribution shifts. T1$\sim$T9 denote different groups of test graph snapshots according to the chronological order. Different groups will have different positive label rates.}
    \label{fig:elliptic}
\end{figure}

\begin{figure}[htp]
    \centering
    \includegraphics[scale=0.4]{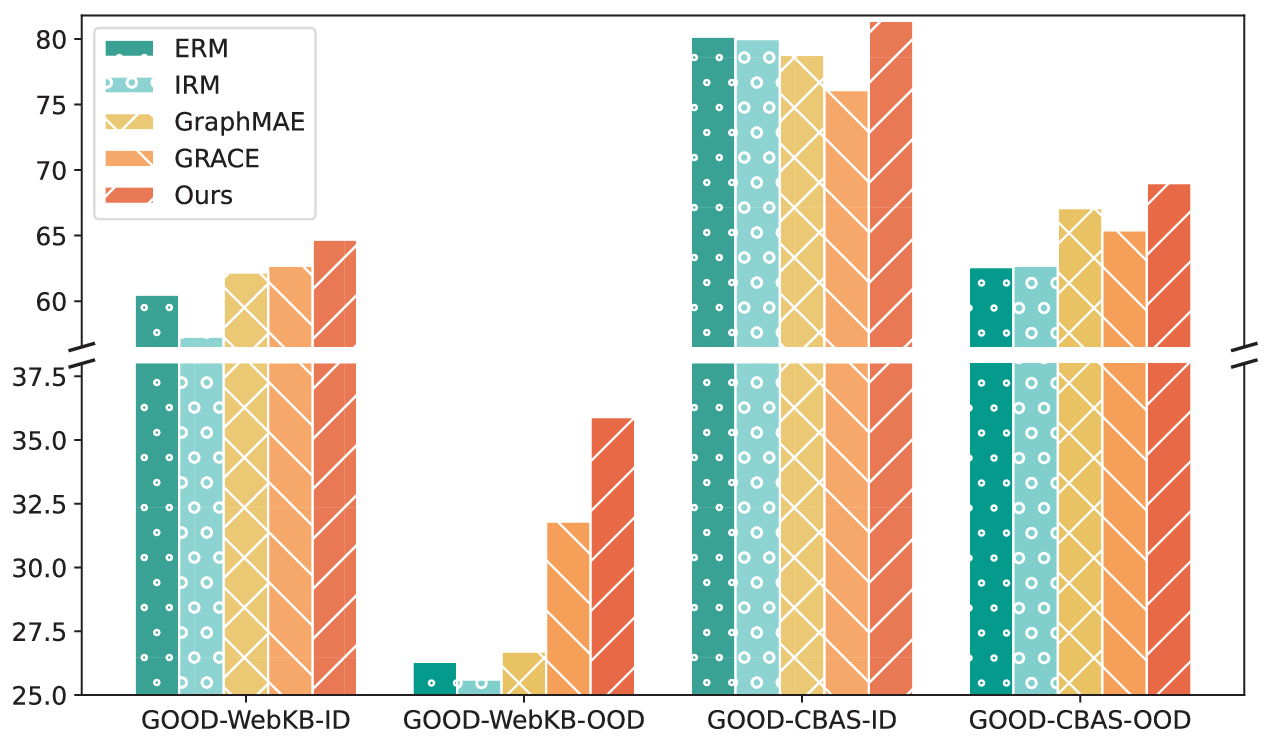}
    \caption{Results on GOOD-WebKB and GOOD-CBAS datasets with concept shift under the inductive setting. `GOOD-WebKB-ID' means in-distribution test performance and `GOOD-WebKB-OOD' means out-of-distribution test performance. So are `GOOD-CBAS-ID' and `GOOD-CBAS-OOD'. We report the mean accuracy across 10 runs.}
    \label{fig:ind_con}
\end{figure}

\begin{figure}[h]
    \centering
    \includegraphics[scale=0.4]{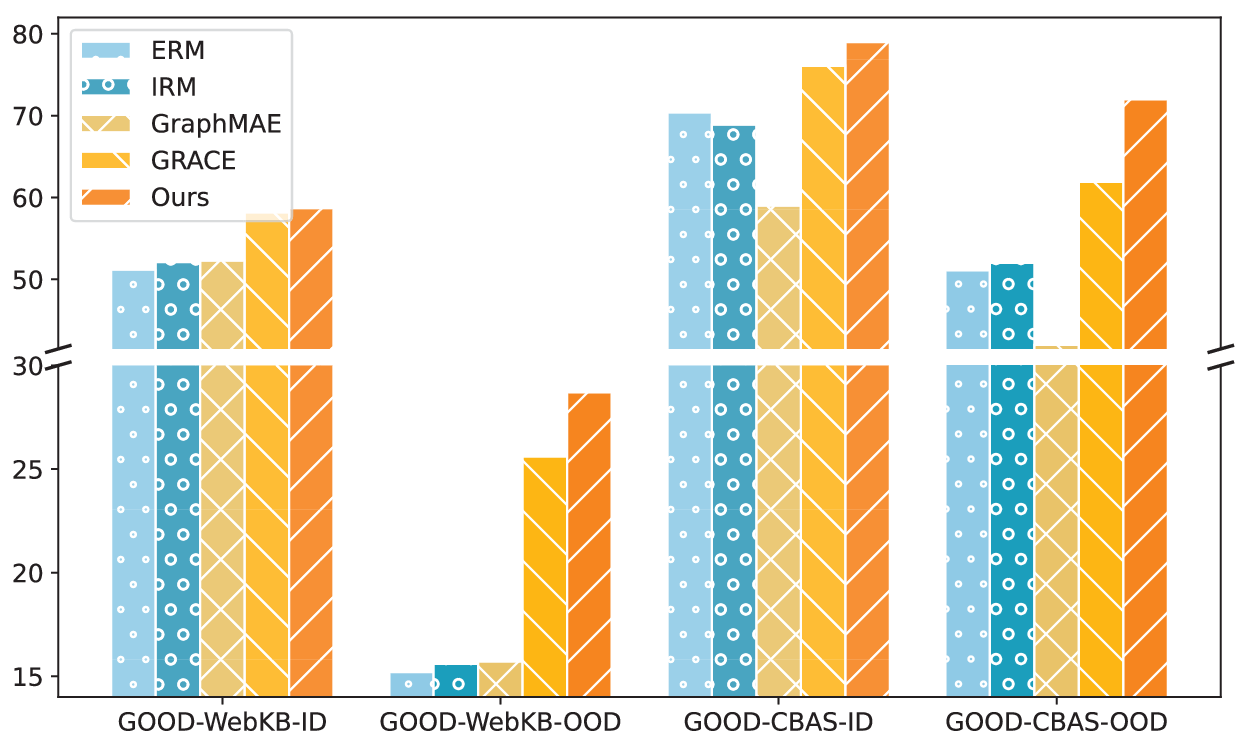}
    \caption{Results on GOOD-WebKB and GOOD-CBAS datasets with covariate shift under the inductive setting.}
    \label{fig:ind_cov}
\end{figure}

\noindent\textbf{Analysis:}
According to Figure~\ref{fig:amazon},\ref{fig:elliptic},\ref{fig:ind_con},\ref{fig:ind_cov}, we can draw following conclusions:
firstly, based on Figure~\ref{fig:amazon}, it is evident that our method outperforms other representative supervised and self-supervised methods on all test graphs (T1$\sim$T8). This superiority is reflected in the larger median value of our method compared to others. For instance, MARIO achieves over a 3\% absolute improvement compared to ERM in terms of the mean value of eight median values. Additionally, our method demonstrates higher stability across different random initializations, as indicated by the closer proximity of the first and third quartile values to the median value~(\eg, the difference of first and third quartile values of ERM, EERM, GRACE and MARIO are 4.2, 3.3, 6.7 and 1.0 on T8 respectively which indicates MARIO is much more stable than other methods). Furthermore, our method exhibits consistent performance across different graphs (\eg, The standard deviation of median values on T1$\sim$T8 for ERM, EERM, GRACE, and MARIO are 0.4, 1.1, 1.2, and 0.3, respectively.), indicating its robustness to environmental variations and its ability to extract invariant features: $g(G^e) \approx g(G^{e'})$ for all $e, e' \in \mathcal{E}^\text{train}$. In summary, our method showcases enhanced OOD generalization capabilities.

Secondly, from the results presented in Figure~\ref{fig:elliptic}, we can observe that our method averagely harvests 10.9\% absolute improvement over GRACE and 12.5\% absolute improvement over EERM in terms of F1 scores on Elliptic dataset. This demonstrates the effectiveness of our method in handling distribution shifts and improving performance compared to existing approaches. It is worth noting that GRACE's performance worsens over time, indicating its inability to handle distribution shifts effectively. In contrast, our method consistently achieves better F1 scores, except for T9, which is caused by the dark market shutdown occurred after T7~\cite{elliptic}. The emergence of such an event introduces significant variations in data distributions, which subsequently results in performance degradation for all methods. Indeed, this event serves as an unpredictable external factor that introduces significant challenges for models trained on limited training data. The results indicate that the performance heavily depends on available training data. Nonetheless, our approach outperforms other methods even in such an extreme case. This highlights the effectiveness of our method in addressing distribution shifts and improving generalization performance.

Finally, based on the observations from Figure~\ref{fig:ind_con} and Figure~\ref{fig:ind_cov} MARIO demonstrates the best performances on both ID and OOD test sets for GOOD-WebKB and GOOD-CBAS datasets, under both concept shift and covariate shift. Notably, MARIO outperforms other methods by more than 3\% and 10\% absolute improvement on GOOD-WebKB and GOOD-CBAS, respectively, under covariate shift. We can draw similar conclusions as discussed in Section~\ref{sec:trans}. Even under the inductive setting, our method continues to demonstrate excellent OOD generalization capabilities and achieves comparable or even improved in-distribution test performance. These statistical results further validate the effectiveness of our method in handling distribution shifts and enhancing generalization performance.

Overall, the observations we have made provide strong evidence of the great capacity of our method for handling distribution shifts, validating its effectiveness and potential for real-world applications.

\subsection{Ablation Studies}\label{sec:ablation}
\noindent Table~\ref{tab:aba} provides a detailed analysis of the effect of each component according to our proposed recipe for improving OOD generalization in graph contrastive learning. Let's examine the different variants of our method and their impact on performance.
Specifically, MARIO~(w/o ad) represents MARIO without  adversarial augmentation. MARIO~(w/o cmi) denotes we only maximize the mutual information between positive pairs without considering conditional mutual information. MARIO~(w/o cmi, ad) means a vanilla graph contrastive method that is similar to GRACE. 

From Table~\ref{tab:aba}, we can find MARIO~(w/o cmi) lags far behind MARIO on OOD test set which demonstrates appropriately minimizing the redundant information (\ie, conditional mutual information) is essential to improve OOD generalization of GCL methods. And adversarial augmentation can also boost OOD generalization because it can approximately serve as a supermum operator to learn more invariant features  discussed in Section~\ref{sec:aug}. Based on the analysis of these variants, it is evident that the proposed improvements on data augmentation and contrastive loss in the recipe are both effective in enhancing graph OOD generalization. Each component contributes to the overall performance improvement, and their combination leads to a stronger self-supervised graph learner in terms of graph OOD generalization. 

In short, the findings from Table~\ref{tab:aba} support the rationale behind your proposed recipe and provide empirical evidence of the effectiveness of each proposed component. By incorporating these enhancements, our method achieves superior performance in handling distribution shifts and improving graph OOD generalization in graph contrastive learning.
\begin{table*}[htp]
\caption{Ablation studies for MARIO by masking each component.}
\label{tab:aba}
\centering
\scalebox{0.9}{
\begin{tabular}{l|cc|cc|cc|cc|cc}
\toprule
\toprule
\multirow{3}{*}{concept shift} & \multicolumn{4}{c|}{GOOD-Cora}                       & \multicolumn{2}{c|}{GOOD-CBAS} & \multicolumn{2}{c|}{GOOD-Twitch} & \multicolumn{2}{c}{GOOD-WebKB} \\
                           & \multicolumn{2}{c}{word} & \multicolumn{2}{c|}{degree}& \multicolumn{2}{c|}{color}    & \multicolumn{2}{c|}{language}   & \multicolumn{2}{c}{university} \\
                           & ID         & OOD         & ID          & OOD          & ID            & OOD           & ID             & OOD            & ID            & OOD            \\
\midrule
MARIO                      & \textbf{67.11±0.46} & \textbf{65.28±0.34}  & \textbf{68.46±0.40}  & \textbf{61.30±0.28}      & \textbf{94.36±1.21}  & \textbf{91.28±1.10}    & 82.31±0.54     & \textbf{63.33±1.72}     & \textbf{65.67±2.81}    & \textbf{37.15±2.37}     \\
MARIO(w/o ad)              & 66.23±0.53 & 64.02±0.18  & 67.88±0.38  & 60.46±0.29   & 93.21±1.25    & 90.29±0.91    & 82.42±0.73     & 60.50±1.02     & 64.83±2.83    & 36.51±3.25    \\
MARIO(w/o cmi)             & 65.32±0.60 & 63.51±0.32  & 68.14±0.32  & 61.19±0.34   & 94.15±1.23    & 90.57±1.96    & \textbf{82.51±0.56}     & 61.41±2.63     & 64.50±4.35    & 35.78±2.53     \\
MARIO(w/o cmi, ad)         & 64.67±0.55 & 63.11±0.32  & 67.95±0.65  & 60.01±0.57   & 93.36±1.66    & 89.64±1.73    & 81.90±0.75     & 60.12±1.60     & 64.17±3.67    & 34.13±2.38     \\
\bottomrule
\end{tabular}}
\end{table*}

\subsection{Sensitivity Analysis}\label{sec:sensitivity}
\begin{figure}
    \centering
    \includegraphics[scale=0.6]{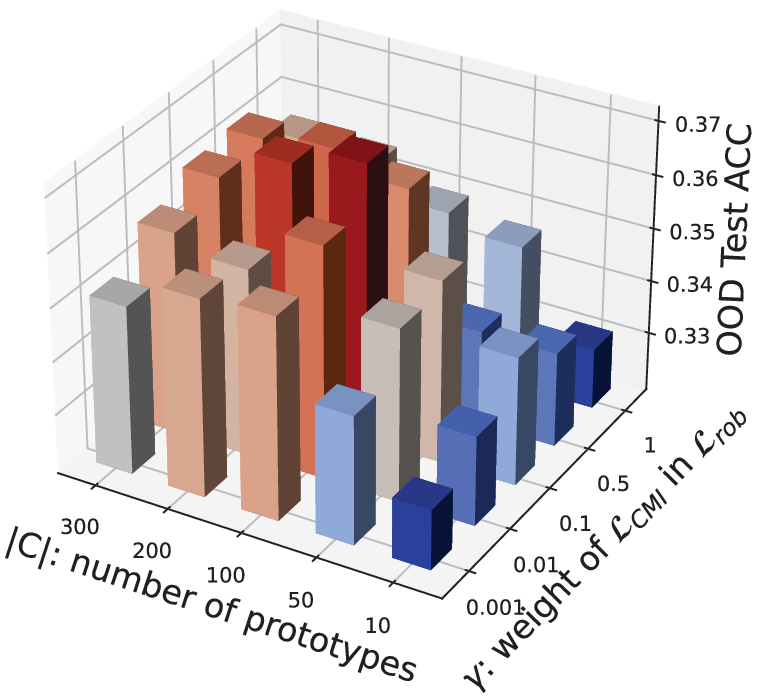}
    \caption{Sensitivity Analysis on CMI coefficient and the number of prototypes.}
    \label{fig:sensitivity}
\end{figure}

\noindent In this subsection, we will analyze some important hyper-parameters of our method. We conduct sensitivity analysis on GOOD-WebKB dataset with concept shift, we chose two sensitive hyper-parameters (\ie, the coefficient $\gamma$ of condition mutual information in Equation~\ref{equ:cmi} and the number of prototypes $|C|$ in Equation~\ref{equ:pq}). The coefficient of CMI range in $[0.001, 0.01, 0.1, 0.5, 1]$ and the number of prototypes $|C|$ ranges in $[10, 50, 100, 200, 300]$. From Figure~\ref{fig:sensitivity}, we can observe that $\gamma$ reaches 0.1 and $|C|$ reaches 100 or 200 can achieve the best OOD test accuracy. Both higher and lower values of $\gamma$ result in suboptimal performance. This finding aligns with previous research such as DIB~\cite{dib}, indicating that an appropriate compression level is crucial for achieving optimal performance. Extremely high or low compression values are not ideal. 

Regarding the number of prototypes $|C|$, based on the results shown in Figure~\ref{fig:sensitivity}, it is found that setting $|C|=100$ leads to the best performance in terms of OOD test accuracy. This choice provides a moderate number of pseudo labels, which is beneficial for the learning process. 

Based on the sensitivity analysis, we determined that setting $\gamma=0.1$ and $|C|=100$ on most datasets. These hyperparameter values strike a balance between compression level and the number of prototypes, resulting in improved graph OOD generalization.

\subsection{Integrated with Other Models}\label{sec:other_models}
\begin{figure*}[hbp]
    \centering
    \includegraphics[scale=0.45]{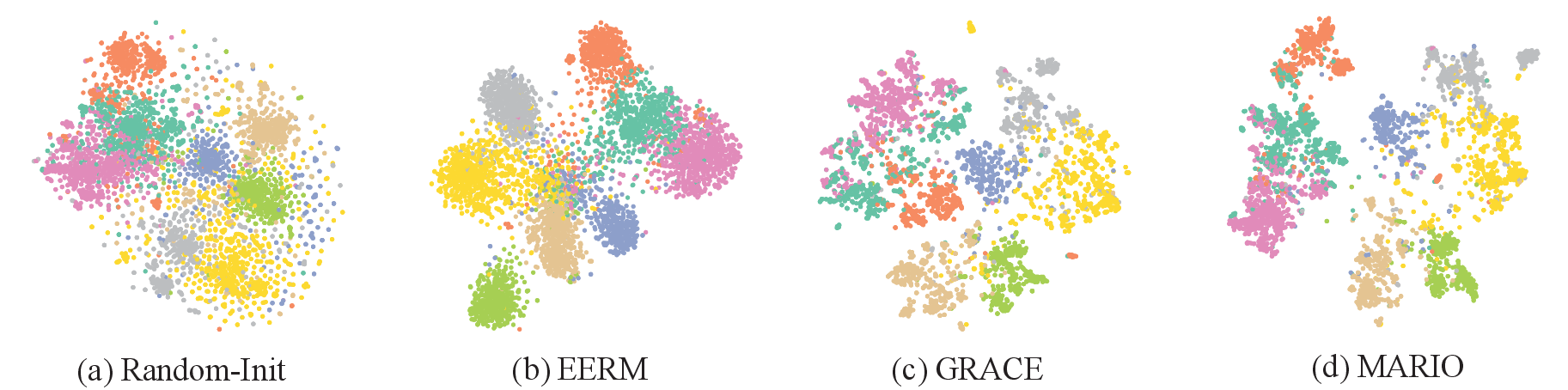} 
    \caption{t-SNE visualization of node embeddings on GOOD-Cora dataset, (a) depicts features from a randomly initialized GCN model, (b) depicts node embeddings from trained EERM, (c) shows embeddings from trained GRACE model, (d) is the result of trained MARIO. The margins of each cluster learned from MARIO are much wider than others.}
    \label{fig:vis}
\end{figure*}

\begin{table}[htp]
\caption{Results of different learning approaches with different encoding models (\ie, GCN, GraphSAGE, GAT).}
\label{tab:others}
\centering
\scalebox{0.9}{
\begin{tabular}{cc|cc|cc}
\toprule
\toprule
\multirow{3}{*}{Model}& \multirow{3}{*}{Method} & \multicolumn{2}{c|}{GOOD-CBAS} & \multicolumn{2}{c}{GOOD-WebKB} \\
                & & \multicolumn{2}{c|}{color}    & \multicolumn{2}{c}{university} \\
                &   & ID          & OOD         & ID          & OOD            \\
\midrule
\multirow{3}{*}{GCN} 
&ERM               & 89.79±1.39 & 83.43±1.19  &  62.67±1.53 & 26.33±1.09         \\
&GRACE             & 92.00±1.39 & 88.64±0.67  &  64.00±3.43 & 34.86±3.43        \\
&MARIO             & 94.36±1.21 & 91.28±1.10  &  65.67±2.81 & 37.15±2.37        \\ \bottomrule
\multirow{3}{*}{SAGE} 
&ERM               & 95.07±1.51 & 75.14±1.19  & 73.67±2.08  & 46.33±3.42       \\
&GRACE             & 95.29±1.11 & 74.43±2.36  & 70.50±5.06  & 49.54±3.83        \\
&MARIO             & 96.00±1.07 & 76.29±3.01  & 71.00±3.82  & 51.74±4.63        \\ \bottomrule
\multirow{3}{*}{GAT} 
&ERM               & 78.64±3.63 & 72.93±2.64  & 61.33±3.71  & 28.99±2.63        \\
&GRACE             & 84.57±1.79 & 78.36±1.60  & 59.50±2.36  & 35.78±3.26        \\
&MARIO             & 84.93±1.95 & 80.43±1.89  & 62.17±4.78  & 38.17±3.10        \\
\bottomrule
\end{tabular}}
\end{table}

\noindent In the subsection, we demonstrate the model-agnostic nature of the recipe by integrating it with various graph neural network (GNN) models, including GCN, GraphSAGE, and GAT.

From Table~\ref{tab:others}, it can be observed that regardless of the specific GNN model used as the encoder, our method consistently achieves the best performance on the OOD test set. This indicates the effectiveness and robustness of our method across different GNN models.
By achieving superior performance across different GNN models, MARIO demonstrates its versatility and ability to improve the OOD generalization of various graph neural models. This highlights the broad applicability and effectiveness of our recipe in enhancing the performance of different GNN encoders.

Furthermore, we integrate our recipe with other GCL methods in Appendix~\ref{app:other_methods}. The results demonstrate our recipe can boost the OOD generalization ability of various GCL methods which means our recipe can serve as a plug-in for many current classical GCL methods.

\begin{figure}[ht]
    \centering
    \includegraphics[scale=0.48]{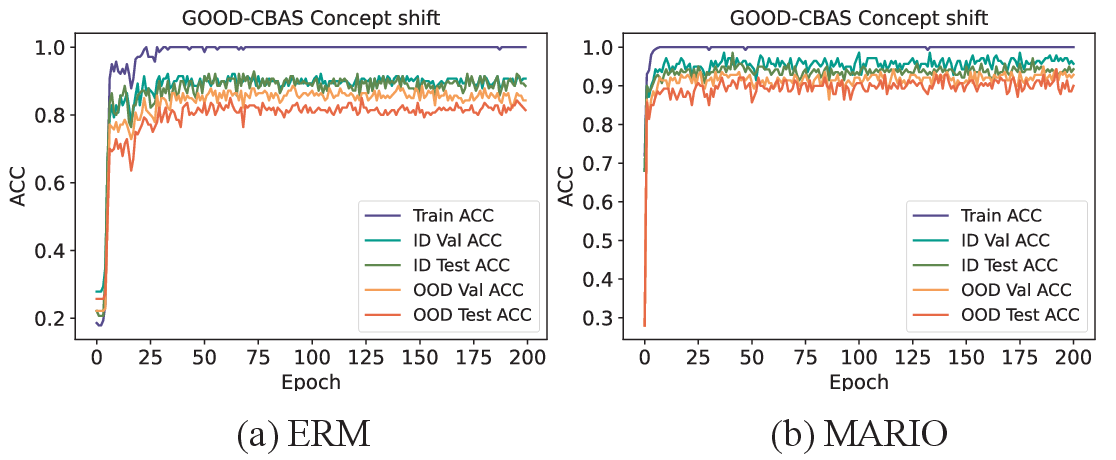}
    \caption{Metric score curves for ERM and MARIO on GOOD-CBAS.}
    \label{fig:curve2}
\end{figure}

\subsection{Visualization}\label{sec:vis}
\subsubsection{Metric Score Curves}
We present metric score curves for ERM and MARIO, including training, ID validation, ID testing, OOD validation, and OOD testing accuracy, in Figure~\ref{fig:curve2}. Notably, MARIO demonstrates superior convergence with approximately 10\% absolute improvement on the OOD test set compared to ERM. Furthermore, MARIO effectively narrows the performance gap between in-distribution and out-of-distribution performance, showcasing its efficacy in enhancing OOD generalization for graph data. More metric score curves can be found in Appendix~\ref{app:curves}.

\subsubsection{Feature Visualization}
In order to assess the quality of learned embeddings, we adopt t-SNE~\cite{tsne} to visualize the node embedding on GOOD-Cora dataset (concept shift in word domain) using random-init of GCN, EERM, GRACE, and MARIO, where different classes have different colors in Figure~\ref{fig:vis}. For clarity, we select eight classes with the largest number of nodes to enhance the informativeness and interpretability of the visualization. We can observe that the 2D projection of node embeddings learned by MARIO has a better separation of clusters, which indicates the model can help learn representative features for downstream tasks. It has to note that we depict both ID nodes and OOD nodes in the same figure. 

Besides, we also separately visualize ID nodes and OOD nodes in the different figures in the Appendix~\ref{app:feature}. And we can find MARIO performs a clearer separation of clusters whether on ID nodes or OOD nodes compared to other methods.

\section{Conclusion}
\noindent In this work, we propose a model-agnostic recipe called MARIO (Model-Agnostic Recipe for Improving OOD Generalization) to address the challenges of distribution shifts in graph contrastive learning. Specifically, this recipe mainly aims to address the drawbacks of the main components (\ie, view generation and representation contrasting) in graph contrastive learning while facing distribution shifts motivated by invariant learning and information bottleneck principles. To the best of our knowledge, this is the first work that investigates the OOD generalization problem of graph contrastive learning specifically for node-level tasks. We conduct substantial experiments to show the superiority of our method on various real-world datasets with diverse distribution shifts. This research contributes to bridging the gap in understanding and addressing distribution shifts in graph contrastive learning, providing valuable insights for future research in this area.

\section*{Acknowledgments}
This work has been supported in part by the NSFC (No. 62272411),  Alibaba-Zhejiang University Joint Research Institute of Frontier Technologies, and Ant Group.
\newpage
\bibliographystyle{IEEEtran}
\bibliography{ref}

\begin{thebibliography}{10}
\providecommand{\url}[1]{#1}
\csname url@samestyle\endcsname
\providecommand{\newblock}{\relax}
\providecommand{\bibinfo}[2]{#2}
\providecommand{\BIBentrySTDinterwordspacing}{\spaceskip=0pt\relax}
\providecommand{\BIBentryALTinterwordstretchfactor}{4}
\providecommand{\BIBentryALTinterwordspacing}{\spaceskip=\fontdimen2\font plus
\BIBentryALTinterwordstretchfactor\fontdimen3\font minus
  \fontdimen4\font\relax}
\providecommand{\BIBforeignlanguage}[2]{{%
\expandafter\ifx\csname l@#1\endcsname\relax
\typeout{** WARNING: IEEEtran.bst: No hyphenation pattern has been}%
\typeout{** loaded for the language `#1'. Using the pattern for}%
\typeout{** the default language instead.}%
\else
\language=\csname l@#1\endcsname
\fi
#2}}
\providecommand{\BIBdecl}{\relax}
\BIBdecl

\bibitem{community}
M.~E. Newman and M.~Girvan, ``Finding and evaluating community structure in
  networks,'' \emph{Physical review E}, vol.~69, no.~2, p. 026113, 2004.

\bibitem{citation}
V.~Batagelj, ``Efficient algorithms for citation network analysis,''
  \emph{arXiv preprint cs/0309023}, 2003.

\bibitem{molecule}
C.~Chen, W.~Ye, Y.~Zuo, C.~Zheng, and S.~P. Ong, ``Graph networks as a
  universal machine learning framework for molecules and crystals,''
  \emph{Chemistry of Materials}, vol.~31, no.~9, pp. 3564--3572, 2019.

\bibitem{gcn}
T.~N. Kipf and M.~Welling, ``Semi-supervised classification with graph
  convolutional networks,'' in \emph{International Conference on Learning
  Representations (ICLR)}, 2017.

\bibitem{sage}
W.~Hamilton, Z.~Ying, and J.~Leskovec, ``Inductive representation learning on
  large graphs,'' \emph{Advances in neural information processing systems},
  vol.~30, 2017.

\bibitem{gin}
\BIBentryALTinterwordspacing
K.~Xu, W.~Hu, J.~Leskovec, and S.~Jegelka, ``How powerful are graph neural
  networks?'' in \emph{International Conference on Learning Representations},
  2019. [Online]. Available: \url{https://openreview.net/forum?id=ryGs6iA5Km}
\BIBentrySTDinterwordspacing

\bibitem{gat}
P.~Veli{\v{c}}kovi{\'c}, G.~Cucurull, A.~Casanova, A.~Romero, P.~Li{\`o}, and
  Y.~Bengio, ``Graph attention networks,'' in \emph{International Conference on
  Learning Representations}.

\bibitem{8294302}
H.~Cai, V.~W. Zheng, and K.~C.-C. Chang, ``A comprehensive survey of graph
  embedding: Problems, techniques, and applications,'' \emph{IEEE Transactions
  on Knowledge and Data Engineering}, vol.~30, no.~9, pp. 1616--1637, 2018.

\bibitem{8519335}
R.~A. Rossi, R.~Zhou, and N.~K. Ahmed, ``Deep inductive graph representation
  learning,'' \emph{IEEE Transactions on Knowledge and Data Engineering},
  vol.~32, no.~3, pp. 438--452, 2020.

\bibitem{erm}
V.~Vapnik, ``Principles of risk minimization for learning theory,''
  \emph{Advances in neural information processing systems}, vol.~4, 1991.

\bibitem{ermxxx}
V.~N. Vapnik, ``An overview of statistical learning theory,'' \emph{IEEE
  transactions on neural networks}, vol.~10, no.~5, pp. 988--999, 1999.

\bibitem{ogb}
W.~Hu, M.~Fey, M.~Zitnik, Y.~Dong, H.~Ren, B.~Liu, M.~Catasta, and J.~Leskovec,
  ``Open graph benchmark: Datasets for machine learning on graphs,''
  \emph{Advances in neural information processing systems}, vol.~33, pp.
  22\,118--22\,133, 2020.

\bibitem{eerm}
Q.~Wu, H.~Zhang, J.~Yan, and D.~Wipf, ``Handling distribution shifts on graphs:
  An invariance perspective,'' in \emph{International Conference on Learning
  Representations}.

\bibitem{good}
S.~Gui, X.~Li, L.~Wang, and S.~Ji, ``Good: A graph out-of-distribution
  benchmark,'' in \emph{Thirty-sixth Conference on Neural Information
  Processing Systems Datasets and Benchmarks Track}.

\bibitem{gil}
H.~Li, Z.~Zhang, X.~Wang, and W.~Zhu, ``Learning invariant graph
  representations for out-of-distribution generalization,'' in \emph{Advances
  in Neural Information Processing Systems}, 2022.

\bibitem{dir}
Y.~Wu, X.~Wang, A.~Zhang, X.~He, and T.-S. Chua, ``Discovering invariant
  rationales for graph neural networks,'' in \emph{International Conference on
  Learning Representations}.

\bibitem{rosa}
\BIBentryALTinterwordspacing
Y.~Zhu, J.~Guo, F.~Wu, and S.~Tang, ``Rosa: A robust self-aligned framework for
  node-node graph contrastive learning,'' in \emph{Proceedings of the
  Thirty-First International Joint Conference on Artificial Intelligence,
  {IJCAI-22}}, L.~D. Raedt, Ed.\hskip 1em plus 0.5em minus 0.4em\relax
  International Joint Conferences on Artificial Intelligence Organization, 7
  2022, pp. 3795--3801, main Track. [Online]. Available:
  \url{https://doi.org/10.24963/ijcai.2022/527}
\BIBentrySTDinterwordspacing

\bibitem{grace}
Y.~Zhu, Y.~Xu, F.~Yu, Q.~Liu, S.~Wu, and L.~Wang, ``{Deep Graph Contrastive
  Representation Learning},'' in \emph{ICML Workshop on Graph Representation
  Learning and Beyond}, 2020.

\bibitem{graphcl}
Y.~You, T.~Chen, Y.~Sui, T.~Chen, Z.~Wang, and Y.~Shen, ``Graph contrastive
  learning with augmentations,'' \emph{Advances in Neural Information
  Processing Systems}, vol.~33, pp. 5812--5823, 2020.

\bibitem{bgrl}
S.~Thakoor, C.~Tallec, M.~G. Azar, R.~Munos, P.~Veli{\v{c}}kovi{\'c}, and
  M.~Valko, ``Bootstrapped representation learning on graphs,'' in \emph{ICLR
  2021 Workshop on Geometrical and Topological Representation Learning}, 2021.

\bibitem{dgi}
P.~Velickovic, W.~Fedus, W.~L. Hamilton, P.~Li{\`{o}}, Y.~Bengio, and R.~D.
  Hjelm, ``Deep graph infomax,'' in \emph{Proc. of ICLR}, 2019.

\bibitem{mvgrl}
K.~Hassani and A.~H. Khasahmadi, ``Contrastive multi-view representation
  learning on graphs,'' in \emph{International Conference on Machine
  Learning}.\hskip 1em plus 0.5em minus 0.4em\relax PMLR, 2020, pp. 4116--4126.

\bibitem{gmae}
\BIBentryALTinterwordspacing
Z.~Hou, X.~Liu, Y.~Cen, Y.~Dong, H.~Yang, C.~Wang, and J.~Tang, ``Graphmae:
  Self-supervised masked graph autoencoders,'' ser. KDD '22.\hskip 1em plus
  0.5em minus 0.4em\relax New York, NY, USA: Association for Computing
  Machinery, 2022, p. 594–604. [Online]. Available:
  \url{https://doi.org/10.1145/3534678.3539321}
\BIBentrySTDinterwordspacing

\bibitem{costa}
Y.~Zhang, H.~Zhu, Z.~Song, P.~Koniusz, and I.~King, ``Costa:
  Covariance-preserving feature augmentation for graph contrastive learning,''
  in \emph{Proceedings of the 28th ACM SIGKDD Conference on Knowledge Discovery
  and Data Mining}, 2022, pp. 2524--2534.

\bibitem{cpc}
A.~v.~d. Oord, Y.~Li, and O.~Vinyals, ``Representation learning with
  contrastive predictive coding,'' \emph{arXiv preprint arXiv:1807.03748},
  2018.

\bibitem{9782500}
J.~Wang, C.~Lan, C.~Liu, Y.~Ouyang, T.~Qin, W.~Lu, Y.~Chen, W.~Zeng, and P.~Yu,
  ``Generalizing to unseen domains: A survey on domain generalization,''
  \emph{IEEE Transactions on Knowledge and Data Engineering}, pp. 1--1, 2022.

\bibitem{8496795}
J.~Lu, A.~Liu, F.~Dong, F.~Gu, J.~Gama, and G.~Zhang, ``Learning under concept
  drift: A review,'' \emph{IEEE Transactions on Knowledge and Data
  Engineering}, vol.~31, no.~12, pp. 2346--2363, 2019.

\bibitem{ood-survey}
Z.~Shen, J.~Liu, Y.~He, X.~Zhang, R.~Xu, H.~Yu, and P.~Cui, ``Towards
  out-of-distribution generalization: A survey,'' \emph{arXiv preprint
  arXiv:2108.13624}, 2021.

\bibitem{good-survey}
H.~Li, X.~Wang, Z.~Zhang, and W.~Zhu, ``Out-of-distribution generalization on
  graphs: A survey,'' \emph{arXiv preprint arXiv:2202.07987}, 2022.

\bibitem{zhu2021shift}
Q.~Zhu, N.~Ponomareva, J.~Han, and B.~Perozzi, ``Shift-robust gnns: Overcoming
  the limitations of localized graph training data,'' \emph{Advances in Neural
  Information Processing Systems}, vol.~34, pp. 27\,965--27\,977, 2021.

\bibitem{wu2020unsupervised}
M.~Wu, S.~Pan, C.~Zhou, X.~Chang, and X.~Zhu, ``Unsupervised domain adaptive
  graph convolutional networks,'' in \emph{Proceedings of The Web Conference
  2020}, 2020, pp. 1457--1467.

\bibitem{zhu2021transfer}
Q.~Zhu, C.~Yang, Y.~Xu, H.~Wang, C.~Zhang, and J.~Han, ``Transfer learning of
  graph neural networks with ego-graph information maximization,''
  \emph{Advances in Neural Information Processing Systems}, vol.~34, pp.
  1766--1779, 2021.

\bibitem{5288526}
S.~J. Pan and Q.~Yang, ``A survey on transfer learning,'' \emph{IEEE
  Transactions on Knowledge and Data Engineering}, vol.~22, no.~10, pp.
  1345--1359, 2010.

\bibitem{groupdro}
S.~Sagawa, P.~W. Koh, T.~B. Hashimoto, and P.~Liang, ``Distributionally robust
  neural networks,'' in \emph{International Conference on Learning
  Representations}.

\bibitem{hu2018does}
W.~Hu, G.~Niu, I.~Sato, and M.~Sugiyama, ``Does distributionally robust
  supervised learning give robust classifiers?'' in \emph{International
  Conference on Machine Learning}.\hskip 1em plus 0.5em minus 0.4em\relax PMLR,
  2018, pp. 2029--2037.

\bibitem{irm}
M.~Arjovsky, L.~Bottou, I.~Gulrajani, and D.~Lopez-Paz, ``Invariant risk
  minimization,'' \emph{arXiv preprint arXiv:1907.02893}, 2019.

\bibitem{sparseirm}
X.~Zhou, Y.~Lin, W.~Zhang, and T.~Zhang, ``Sparse invariant risk
  minimization,'' in \emph{International Conference on Machine Learning}.\hskip
  1em plus 0.5em minus 0.4em\relax PMLR, 2022, pp. 27\,222--27\,244.

\bibitem{peters2016causal}
J.~Peters, P.~B{\"u}hlmann, and N.~Meinshausen, ``Causal inference by using
  invariant prediction: identification and confidence intervals,''
  \emph{Journal of the Royal Statistical Society Series B: Statistical
  Methodology}, vol.~78, no.~5, pp. 947--1012, 2016.

\bibitem{heinze2018causal}
C.~Heinze-Deml, J.~Peters, and N.~Meinshausen, ``Invariant causal prediction
  for nonlinear models,'' \emph{Journal of Causal Inference}, vol.~6, no.~2, p.
  20170016, 2018.

\bibitem{gsat}
S.~Miao, M.~Liu, and P.~Li, ``Interpretable and generalizable graph learning
  via stochastic attention mechanism,'' in \emph{International Conference on
  Machine Learning}.\hskip 1em plus 0.5em minus 0.4em\relax PMLR, 2022, pp.
  15\,524--15\,543.

\bibitem{dib}
A.~A. Alemi, I.~Fischer, J.~V. Dillon, and K.~Murphy, ``Deep variational
  information bottleneck,'' \emph{arXiv preprint arXiv:1612.00410}, 2016.

\bibitem{9764632}
Y.~Xie, Z.~Xu, J.~Zhang, Z.~Wang, and S.~Ji, ``Self-supervised learning of
  graph neural networks: A unified review,'' \emph{IEEE Transactions on Pattern
  Analysis and Machine Intelligence}, vol.~45, no.~2, pp. 2412--2429, 2023.

\bibitem{wang2020understanding}
T.~Wang and P.~Isola, ``Understanding contrastive representation learning
  through alignment and uniformity on the hypersphere,'' in \emph{International
  Conference on Machine Learning}.\hskip 1em plus 0.5em minus 0.4em\relax PMLR,
  2020, pp. 9929--9939.

\bibitem{zhu2023sgl}
Y.~Zhu, J.~Guo, and S.~Tang, ``Sgl-pt: A strong graph learner with graph prompt
  tuning,'' \emph{arXiv preprint arXiv:2302.12449}, 2023.

\bibitem{zhang2023structure}
Z.~Zhang, Y.~Zhu, H.~Shi, and S.~Tang, ``Structure-aware group discrimination
  with adaptive-view graph encoder: A fast graph contrastive learning
  framework,'' \emph{arXiv preprint arXiv:2303.05231}, 2023.

\bibitem{raft}
H.~Shi, D.~Luo, S.~Tang, J.~Wang, and Y.~Zhuang, ``Run away from your teacher:
  Understanding byol by a novel self-supervised approach,'' \emph{arXiv
  preprint arXiv:2011.10944}, 2020.

\bibitem{9770382}
Y.~Liu, M.~Jin, S.~Pan, C.~Zhou, Y.~Zheng, F.~Xia, and P.~S. Yu, ``Graph
  self-supervised learning: A survey,'' \emph{IEEE Transactions on Knowledge
  and Data Engineering}, vol.~35, no.~6, pp. 5879--5900, 2023.

\bibitem{rgcl}
S.~Li, X.~Wang, A.~Zhang, Y.~Wu, X.~He, and T.-S. Chua, ``Let invariant
  rationale discovery inspire graph contrastive learning,'' in
  \emph{International Conference on Machine Learning}.\hskip 1em plus 0.5em
  minus 0.4em\relax PMLR, 2022, pp. 13\,052--13\,065.

\bibitem{arcl}
X.~Zhao, T.~Du, Y.~Wang, J.~Yao, and W.~Huang, ``Arcl: Enhancing contrastive
  learning with augmentation-robust representations,'' in \emph{The Eleventh
  International Conference on Learning Representations}.

\bibitem{huang2023towards}
\BIBentryALTinterwordspacing
W.~Huang, M.~Yi, X.~Zhao, and Z.~Jiang, ``Towards the generalization of
  contrastive self-supervised learning,'' in \emph{The Eleventh International
  Conference on Learning Representations}, 2023. [Online]. Available:
  \url{https://openreview.net/forum?id=XDJwuEYHhme}
\BIBentrySTDinterwordspacing

\bibitem{shi2022robust}
Y.~Shi, I.~Daunhawer, J.~E. Vogt, P.~H. Torr, and A.~Sanyal, ``How robust are
  pre-trained models to distribution shift?'' \emph{arXiv preprint
  arXiv:2206.08871}, 2022.

\bibitem{simclr}
T.~Chen, S.~Kornblith, M.~Norouzi, and G.~Hinton, ``A simple framework for
  contrastive learning of visual representations,'' in \emph{International
  conference on machine learning}.\hskip 1em plus 0.5em minus 0.4em\relax PMLR,
  2020, pp. 1597--1607.

\bibitem{tian2020makes}
Y.~Tian, C.~Sun, B.~Poole, D.~Krishnan, C.~Schmid, and P.~Isola, ``What makes
  for good views for contrastive learning?'' \emph{Advances in neural
  information processing systems}, vol.~33, pp. 6827--6839, 2020.

\bibitem{ahuja2021invariance}
K.~Ahuja, E.~Caballero, D.~Zhang, J.-C. Gagnon-Audet, Y.~Bengio, I.~Mitliagkas,
  and I.~Rish, ``Invariance principle meets information bottleneck for
  out-of-distribution generalization,'' \emph{Advances in Neural Information
  Processing Systems}, vol.~34, pp. 3438--3450, 2021.

\bibitem{li2022invariant}
B.~Li, Y.~Shen, Y.~Wang, W.~Zhu, D.~Li, K.~Keutzer, and H.~Zhao, ``Invariant
  information bottleneck for domain generalization,'' in \emph{Proceedings of
  the AAAI Conference on Artificial Intelligence}, vol.~36, no.~7, 2022, pp.
  7399--7407.

\bibitem{shafahi2019adversarial}
A.~Shafahi, M.~Najibi, M.~A. Ghiasi, Z.~Xu, J.~Dickerson, C.~Studer, L.~S.
  Davis, G.~Taylor, and T.~Goldstein, ``Adversarial training for free!''
  \emph{Advances in Neural Information Processing Systems}, vol.~32, 2019.

\bibitem{flag}
K.~Kong, G.~Li, M.~Ding, Z.~Wu, C.~Zhu, B.~Ghanem, G.~Taylor, and T.~Goldstein,
  ``Robust optimization as data augmentation for large-scale graphs,'' in
  \emph{Proceedings of the IEEE/CVF Conference on Computer Vision and Pattern
  Recognition}, 2022, pp. 60--69.

\bibitem{kim2020adversarial}
M.~Kim, J.~Tack, and S.~J. Hwang, ``Adversarial self-supervised contrastive
  learning,'' \emph{Advances in Neural Information Processing Systems},
  vol.~33, pp. 2983--2994, 2020.

\bibitem{suresh2021adversarial}
S.~Suresh, P.~Li, C.~Hao, and J.~Neville, ``Adversarial graph augmentation to
  improve graph contrastive learning,'' \emph{Advances in Neural Information
  Processing Systems}, vol.~34, pp. 15\,920--15\,933, 2021.

\bibitem{jiang2020robust}
Z.~Jiang, T.~Chen, T.~Chen, and Z.~Wang, ``Robust pre-training by adversarial
  contrastive learning,'' \emph{Advances in neural information processing
  systems}, vol.~33, pp. 16\,199--16\,210, 2020.

\bibitem{tishby_ib}
N.~Tishby and N.~Zaslavsky, ``Deep learning and the information bottleneck
  principle,'' in \emph{2015 ieee information theory workshop (itw)}.\hskip 1em
  plus 0.5em minus 0.4em\relax IEEE, 2015, pp. 1--5.

\bibitem{gib}
T.~Wu, H.~Ren, P.~Li, and J.~Leskovec, ``Graph information bottleneck,''
  \emph{Advances in Neural Information Processing Systems}, vol.~33, pp.
  20\,437--20\,448, 2020.

\bibitem{donsker1975asymptotic}
M.~D. Donsker and S.~S. Varadhan, ``Asymptotic evaluation of certain markov
  process expectations for large time, i,'' \emph{Communications on Pure and
  Applied Mathematics}, vol.~28, no.~1, pp. 1--47, 1975.

\bibitem{mine}
M.~I. Belghazi, A.~Baratin, S.~Rajeshwar, S.~Ozair, Y.~Bengio, A.~Courville,
  and D.~Hjelm, ``Mutual information neural estimation,'' in
  \emph{International conference on machine learning}.\hskip 1em plus 0.5em
  minus 0.4em\relax PMLR, 2018, pp. 531--540.

\bibitem{fgan}
S.~Nowozin, B.~Cseke, and R.~Tomioka, ``f-gan: Training generative neural
  samplers using variational divergence minimization,'' \emph{Advances in
  neural information processing systems}, vol.~29, 2016.

\bibitem{nce}
M.~Gutmann and A.~Hyv{\"a}rinen, ``Noise-contrastive estimation: A new
  estimation principle for unnormalized statistical models,'' in
  \emph{Proceedings of the thirteenth international conference on artificial
  intelligence and statistics}.\hskip 1em plus 0.5em minus 0.4em\relax JMLR
  Workshop and Conference Proceedings, 2010, pp. 297--304.

\bibitem{pcl}
J.~Li, P.~Zhou, C.~Xiong, and S.~Hoi, ``Prototypical contrastive learning of
  unsupervised representations,'' in \emph{International Conference on Learning
  Representations}.

\bibitem{swav}
M.~Caron, I.~Misra, J.~Mairal, P.~Goyal, P.~Bojanowski, and A.~Joulin,
  ``Unsupervised learning of visual features by contrasting cluster
  assignments,'' \emph{Advances in neural information processing systems},
  vol.~33, pp. 9912--9924, 2020.

\bibitem{bilevel}
R.~Liu, J.~Gao, J.~Zhang, D.~Meng, and Z.~Lin, ``Investigating bi-level
  optimization for learning and vision from a unified perspective: A survey and
  beyond,'' \emph{IEEE Transactions on Pattern Analysis and Machine
  Intelligence}, vol.~44, no.~12, pp. 10\,045--10\,067, 2021.

\bibitem{ji2022drugood}
Y.~Ji, L.~Zhang, J.~Wu, B.~Wu, L.-K. Huang, T.~Xu, Y.~Rong, L.~Li, J.~Ren,
  D.~Xue \emph{et~al.}, ``Drugood: Out-of-distribution (ood) dataset curator
  and benchmark for ai-aided drug discovery--a focus on affinity prediction
  problems with noise annotations,'' \emph{arXiv preprint arXiv:2201.09637},
  2022.

\bibitem{gds}
\BIBentryALTinterwordspacing
M.~Ding, K.~Kong, J.~Chen, J.~Kirchenbauer, M.~Goldblum, D.~Wipf, F.~Huang, and
  T.~Goldstein, ``A closer look at distribution shifts and out-of-distribution
  generalization on graphs,'' in \emph{NeurIPS 2021 Workshop on Distribution
  Shifts: Connecting Methods and Applications}, 2021. [Online]. Available:
  \url{https://openreview.net/forum?id=XvgPGWazqRH}
\BIBentrySTDinterwordspacing

\bibitem{yang2016revisiting}
Z.~Yang, W.~Cohen, and R.~Salakhudinov, ``Revisiting semi-supervised learning
  with graph embeddings,'' in \emph{International conference on machine
  learning}.\hskip 1em plus 0.5em minus 0.4em\relax PMLR, 2016, pp. 40--48.

\bibitem{gilmer2017neural}
J.~Gilmer, S.~S. Schoenholz, P.~F. Riley, O.~Vinyals, and G.~E. Dahl, ``Neural
  message passing for quantum chemistry,'' in \emph{International conference on
  machine learning}.\hskip 1em plus 0.5em minus 0.4em\relax PMLR, 2017, pp.
  1263--1272.

\bibitem{gae}
T.~N. Kipf and M.~Welling, ``Variational graph auto-encoders,'' \emph{NIPS
  Workshop on Bayesian Deep Learning}, 2016.

\bibitem{ahuja2020empirical}
K.~Ahuja, J.~Wang, A.~Dhurandhar, K.~Shanmugam, and K.~R. Varshney, ``Empirical
  or invariant risk minimization? a sample complexity perspective,''
  \emph{arXiv preprint arXiv:2010.16412}, 2020.

\bibitem{rosenfeld2021risks}
E.~Rosenfeld, P.~Ravikumar, and A.~Risteski, ``The risks of invariant risk
  minimization,'' in \emph{International Conference on Learning
  Representations}, vol.~9, 2021.

\bibitem{elliptic}
A.~Pareja, G.~Domeniconi, J.~Chen, T.~Ma, T.~Suzumura, H.~Kanezashi, T.~Kaler,
  T.~Schardl, and C.~Leiserson, ``Evolvegcn: Evolving graph convolutional
  networks for dynamic graphs,'' in \emph{Proceedings of the AAAI conference on
  artificial intelligence}, vol.~34, no.~04, 2020, pp. 5363--5370.

\bibitem{tsne}
L.~Van~der Maaten and G.~Hinton, ``Visualizing data using t-sne.''
  \emph{Journal of machine learning research}, vol.~9, no.~11, 2008.

\bibitem{ma2021conditional}
M.~Q. Ma, Y.-H.~H. Tsai, P.~P. Liang, H.~Zhao, K.~Zhang, R.~Salakhutdinov, and
  L.-P. Morency, ``Conditional contrastive learning for improving fairness in
  self-supervised learning,'' \emph{arXiv preprint arXiv:2106.02866}, 2021.

\bibitem{nguyen2010estimating}
X.~Nguyen, M.~J. Wainwright, and M.~I. Jordan, ``Estimating divergence
  functionals and the likelihood ratio by convex risk minimization,''
  \emph{IEEE Transactions on Information Theory}, vol.~56, no.~11, pp.
  5847--5861, 2010.

\bibitem{cuturi2013sinkhorn}
M.~Cuturi, ``Sinkhorn distances: Lightspeed computation of optimal transport,''
  \emph{Advances in neural information processing systems}, vol.~26, 2013.

\bibitem{byol}
J.-B. Grill, F.~Strub, F.~Altch{\'e}, C.~Tallec, P.~Richemond, E.~Buchatskaya,
  C.~Doersch, B.~Avila~Pires, Z.~Guo, M.~Gheshlaghi~Azar \emph{et~al.},
  ``Bootstrap your own latent-a new approach to self-supervised learning,''
  \emph{Advances in neural information processing systems}, vol.~33, pp.
  21\,271--21\,284, 2020.

\bibitem{bojchevskideep}
A.~Bojchevski and S.~G{\"u}nnemann, ``Deep gaussian embedding of graphs:
  Unsupervised inductive learning via ranking,'' in \emph{International
  Conference on Learning Representations}.

\bibitem{ying2019gnnexplainer}
Z.~Ying, D.~Bourgeois, J.~You, M.~Zitnik, and J.~Leskovec, ``Gnnexplainer:
  Generating explanations for graph neural networks,'' \emph{Advances in neural
  information processing systems}, vol.~32, 2019.

\bibitem{moco}
K.~He, H.~Fan, Y.~Wu, S.~Xie, and R.~Girshick, ``Momentum contrast for
  unsupervised visual representation learning,'' in \emph{Proceedings of the
  IEEE/CVF conference on computer vision and pattern recognition}, 2020, pp.
  9729--9738.

\bibitem{fey2019fast}
M.~Fey and J.~E. Lenssen, ``Fast graph representation learning with pytorch
  geometric,'' \emph{ArXiv preprint}, 2019.

\bibitem{torch}
A.~Paszke, S.~Gross, F.~Massa, A.~Lerer, J.~Bradbury, G.~Chanan, T.~Killeen,
  Z.~Lin, N.~Gimelshein, L.~Antiga, A.~Desmaison, A.~K{\"{o}}pf, E.~Yang,
  Z.~DeVito, M.~Raison, A.~Tejani, S.~Chilamkurthy, B.~Steiner, L.~Fang,
  J.~Bai, and S.~Chintala, ``Pytorch: An imperative style, high-performance
  deep learning library,'' in \emph{Proc. of NeurIPS}, 2019.

\bibitem{pedregosa2011scikit}
F.~Pedregosa, G.~Varoquaux, A.~Gramfort, V.~Michel, B.~Thirion, O.~Grisel,
  M.~Blondel, P.~Prettenhofer, R.~Weiss, V.~Dubourg \emph{et~al.},
  ``Scikit-learn: Machine learning in python,'' \emph{the Journal of machine
  Learning research}, vol.~12, pp. 2825--2830, 2011.

\end{thebibliography}

\newpage
~
\newpage
\appendices{
\section{Proofs in Section~\ref{sec:pre}}\label{app:proof_recipe1}
In this section, we will illustrate some notations mentioned in Equation~\ref{equ:cl_down} firstly, and then we will provide a proof of Equation~\ref{equ:cl_down}.
\begin{definition}[$(\sigma,\delta)$-augmentation\cite{huang2023towards}] Let $C_k \subseteq \mathcal{G}$ denote the set of all points in class $k$. A graph augmentation set $\mathcal{T}$ can be referred to as a $(\sigma,\delta)$-augmentation on $\mathcal{G}$, where $\sigma \in (0,1]$ and $\delta > 0$. This is the case if, for every $k \in [K]$, there exists a subset $C_k^0 \subseteq C_k$ such that the following conditions hold:

\begin{equation}
\begin{aligned}
& P_{G \sim \mathcal{G}}\left(G \in C_k^0\right) \geq \sigma {P}_{G \sim \mathcal{G}}\left(G \in C_k\right), \\ 
& \text {and } \sup _{G_1, G_2 \in C_k^0} d_A\left(G_1, G_2\right) \leq \delta , \\
\end{aligned}
\end{equation}
where $d_{\mathcal{A}}\left(G_1, G_2\right):=\inf _{\tau_1, \tau_2 \in \mathcal{T}} d\left(\tau_1\left(G\right), \tau_2\left(G\right)\right)$ for some distance $d(\cdot, \cdot)$.

\end{definition}
This definition quantifies the concentration of augmented data. An augmentation set with a smaller value of $\delta$ and a larger value of $\sigma$ results in a more clustered arrangement of the original data. In other words, samples from the same class are closer to each other after augmentation. Consequently, one can anticipate that the learned representation $g_\theta$ will exhibit improved cluster performance. This principle was proposed in \cite{huang2023towards} and modified to a more practical scenario by \cite{arcl}.

\noindent\textbf{Proof of Equation~\ref{equ:cl_down}.} For simplicity, we omit the notations $\mathcal{G}$ and $\pi$ here and omit the parameters $\theta,\omega$ of $g_\theta$ and $p_\omega$, and use $G_1$ to denote the augmented data, use $t_k$ to represent $\mathcal{G}_{\pi}(C_k)$ and use $\mu_k$ to denote $\mu_k(g;\mathcal{G}_{\pi})$. Based on Theorem 2, Lemma B.1 in \cite{huang2023towards}, and Appendix B in \cite{arcl}, we have
\begin{equation}
\underset{G \in C_k}{\mathbb{E}}\left\|g\left(G_1\right)-\mu_k\right\| \leq c \sqrt{\frac{1}{t_k}} \mathcal{L}_{\text {align }}^{\frac{1}{4}}(g)+\zeta(\sigma, \delta)
\end{equation}
for some constant $c$, where
\begin{equation}
\zeta (\sigma, \delta):=4\left(1-\sigma\left(1-\frac{L \delta}{4}\right)\right)
\end{equation}
We can find $\zeta$ is decreasing with $\sigma$ and increasing with $\delta$. So, we can obtain:
\begin{equation}
\begin{aligned}
& c\left(\sum_{k=1}^K \sqrt{t_k}\right) \mathcal{L}_{\mathrm{align}}^{\frac{1}{4}}(f)+\zeta(\sigma, \delta) \\
& \geq \sum_{k=1}^K t_k \underset{G \in C_k}{\mathbb{E}}\left\|g\left(G_1\right)-\mu_k\right\| \\
& \geq \sum_{k=1}^K \frac{t_k}{\|p\|} \underset{G \in C_k}{\mathbb{E}} \underset{G_1 \in \tau(G)}{\mathbb{E}}\left\|p \circ g\left(G_1\right)-p \circ \mu_k\right\| \\
& \geq \sum_{k=1}^K \frac{t_k}{\|p\|} \underset{G \in C_k}{\mathbb{E}} \underset{G_1  \in \tau(G)}{\mathbb{E}}\left\|p \circ g\left(G_1\right)-e_k\right\| \\ 
&\quad\quad\quad-\frac{1}{\|p\|} \sum_{k=1}^K t_k\left\|e_k-p \circ \mu_k\right\| \\
& =\frac{1}{\|p\|} \mathcal{R}(p \circ g)-\frac{1}{\|p\|} \sum_{k=1}^K t_k\left\|e_k-p \circ \mu_k\right\|
\end{aligned}
\end{equation}
for all linear layer $p \in \mathcal{R}^{K \times d_1}$. Therefore, we obtain
\begin{equation}
\begin{aligned}
\mathcal{R}(p \circ g) & \leq c\|p\| \mathcal{L}_{\text {align }}^{\frac{1}{4}} \sum_{k=1}^K \sqrt{t_k}+\|p\| \zeta(\sigma, \delta)\\
&\quad\quad+\sum_{k=1}^K t_k\left\|e_k-p \circ \mu_k\right\| \\
& \leq c\|p\| \sqrt{K} \mathcal{L}_{\text {align}}^{\frac{1}{4}}+\|p\| \zeta(\sigma, \delta) \\ 
& \quad\quad +\sum_{k=1}^K t_k\left\|e_k-p \circ \mu_k\right\|.
\end{aligned}
\end{equation}




\section{Special Case\label{app:case}}
The case adopted from~\cite{arcl} is used to prove the encoders learned through contrastive learning could behave extremely differently in different $\mathcal{G}_{\tau}$.
\begin{proposition}
    Consider a binary classification problem with data $(X_1, X_2) \sim \mathcal{N}(0, I_2)$. If $X_1 \geq 0$, the label $Y=1$, and the data augmentation is to multiply $X_2$ by standard normal noise:
    \begin{equation}
\begin{aligned}
\tau_\theta(X) & =\left(X_1, \theta \cdot X_2\right) \\
\theta & \sim \mathcal{N}(0,1)
\end{aligned}
\end{equation}
The transformation-induced domain set is $\mathcal{B}=\left\{\mathcal{G}_c: \mathcal{G}_c=\left(X_1, c \cdot X_2\right) \text { for } c \in \mathbb{R}\right\}$. Considering the 0-1 loss, $\forall \varepsilon \ge 0$, there holds representation $g$ and two domains $\mathcal{G}_c$ and $\mathcal{G}_{c^\prime}$ such that
\begin{equation}
\mathcal{L}_{\text {align }}(g ; \mathcal{G}, \pi)<\varepsilon
\end{equation}
but $g$ behaves extremely differently in different domains  $\mathcal{G}_c$ and $\mathcal{G}_{c^\prime}$:
\begin{equation}
\left|\mathcal{R}\left(g ; \mathcal{G}_c\right)-\mathcal{R}\left(g ; \mathcal{G}_{c^{\prime}}\right)\right| \geq \frac{1}{4}
\end{equation}
This instance\footnote{For simplicity, we assume the adjacent matrix is an identity matrix here.} illustrates that the obtained representation with small contrastive loss will still exhibit significantly varied performance over different augmentation-induced domains. The underlying idea behind this example lies in achieving a small $\mathcal{L}_{\text {align}}$ by aligning different augmentation-induced domains in an average sense, rather than a uniform one. Consequently, the representation might still encounter large alignment losses on certain infrequently chosen augmented domains.
\end{proposition}
\emph{Proof.} For $\varepsilon \ge 0$, let $t=\sqrt{\varepsilon}/2$ and $g(x_1, x_2)=x1+tx_2.$ Then, the alignment loss of $g$ satisfies:
\begin{equation}
\mathcal{L}_{\text {align }}(g ; \mathcal{G}, \pi)=t^2 \mathbb{E} X_2^2 \underset{\left(\theta_1, \theta_2\right) \sim \mathcal{N}(0,1)^2}{\mathbb{E}}\left(\theta_1-\theta_2\right)^2=2 t^2<\varepsilon .
\end{equation}
Set $c$ as 0 and $c^{\prime}$ as $1/t$, it is obviously that:
\begin{equation}
\mathcal{R}\left(g ; \mathcal{G}_c\right)=0
\end{equation}
but
\begin{equation}
\begin{aligned}
&\mathcal{R}\left(g ; \mathcal{G}_{c^{\prime}}\right)= \\ & P\left(X_1<0, X_1+X_2 \geq 0\right)+P\left(X_1 \geq 0, X_1+X_2 \leq 0\right)=\frac{1}{4}
\end{aligned}
\end{equation}
\section{Proof of Theorem~\ref{theorem:align}}\label{app:align}
\emph{Proof of Theorem~\ref{theorem:align}}:
\begin{equation}
\begin{aligned}
& \mathcal{R}\left(p \circ g ; \mathcal{G}_\tau\right)-\mathcal{R}\left(p \circ g ; \mathcal{G}_{\tau^{\prime}}\right) \\ 
& \quad =\underset{(G, Y) \sim \mathcal{G}}{\mathbb{E}}\left(|p \circ g(\tau(G))-Y|^2-\left|p \circ g\left(\tau^{\prime} (G)\right)-Y\right|^2\right) \\
& \quad =\underset{(G, Y) \sim \mathcal{G}}{\mathbb{E}}\left(p \circ g(\tau( G))-p \circ g\left(\tau^{\prime} (G)\right)\right)\left(\left(p \circ g(\tau (G))+ \right.\right.\\  & \quad\quad\quad\quad\quad\quad\quad\quad\quad\quad\quad\quad\quad\quad\quad\quad \left.\left. p \circ g\left(\tau^{\prime} (G)\right)\right)+2 Y\right) \\
& \quad \leq c \underset{(G, Y) \sim \mathcal{G}}{\mathbb{E}}\left\|p \circ g(\tau (G))-p \circ g\left(\tau^{\prime} (G)\right)\right\| \\
& \quad \leq c\|p\| \underset{(G, Y) \sim \mathcal{G}}{\mathbb{E}}\left\|g(\tau (G))-g\left(\tau^{\prime} (G)\right)\right\| \\
& \quad \leq c\|p\| \mathcal{L}_{\text{align}^*}(g)
\end{aligned}
\end{equation}

\section{Proof of CMI Lower Bound\label{app:cmi}}
In this section, we provide a theoretical justification of why Equation~\ref{equ:con} is a lower bound of CMI. And some justifications are borrowed from~\cite{ma2021conditional,nguyen2010estimating}. Firstly, we present the following lemmas which will be used. 
\subsection{Fundamental Lemmas}
\begin{lemma}
    Let $U$ and $V$ be two random variables whose sample spaces are $\mathcal{U}$ and $\mathcal{V}$, $f: (\mathcal{U}\times\mathcal{V}) \rightarrow \mathbb{R}$ be a mapping function, and $\mathbb{P}$ and $\mathbb{Q}$ be the probability measures on $\mathcal{U}\times\mathcal{V}$, we can obtain:
\begin{equation}
D_{\mathrm{KL}}(\mathbb{P} \| \mathbb{Q})=\sup _f \mathbb{E}_{(u, v) \sim \mathbb{P}}[f(u, v)]-\mathbb{E}_{(u, v) \sim \mathbb{Q}}\left[e^{f(u, v)}\right]+1
\label{equ:kl}
\end{equation}
\label{lem:kl}
\end{lemma}
\emph{Proof.} The second-order functional derivative of the above function is $-e^{f(u, v)} \cdot d \mathbb{Q}$. This negative term means Equation~\ref{equ:kl} has a supreme value. Through setting the first-order functional derivative as zero $d \mathbb{P}-e^{f(u, v)} \cdot d \mathbb{Q}=0$, we can get the optimal mapping function $f^*(u, v)=\log \frac{d \mathbb{P}}{d \mathbb{Q}}$. Rewrite the Equation~\ref{equ:kl} with $f^*$:
\begin{equation}
\mathbb{E}_{\mathbb{P}}\left[f^*(u, v)\right]-\mathbb{E}_{\mathbb{Q}}\left[e^{f^*(u, v)}\right]+1=\mathbb{E}_{\mathbb{P}}\left[\log \frac{d \mathbb{P}}{d \mathbb{Q}}\right]=D_{\mathrm{KL}}(\mathbb{P} \| \mathbb{Q})
\end{equation}

\begin{lemma}
    Let $U$, $V$, and $Y$ be three random variables whose sample spaces are $\mathcal{U}$, $\mathcal{V}$ and $\mathcal{Y}$, $f: (\mathcal{U}\times\mathcal{V} \times \mathcal{Y}) \rightarrow \mathbb{R}$ be a mapping function, and $\mathbb{P}$ and $\mathbb{Q}$ be the probability measures on $\mathcal{U}\times\mathcal{V}\times\mathcal{Y}$, we can obtain:
\begin{equation}
\begin{aligned}
    D_{\mathrm{KL}}(\mathbb{P} \| \mathbb{Q})=\sup _f & \mathbb{E}_{(u, v, y) \sim \mathbb{P}}[f(u, v, y)]\\
                                                      & -\mathbb{E}_{(u, v, y) \sim \mathbb{Q}}\left[e^{f(u, v, y)}\right]+1
\end{aligned}
\label{equ:kl2}
\end{equation}
\label{lem:kl2}
\end{lemma}
\emph{Proof.} The second-order functional derivative of the above function is $-e^{f(u, v, y)} \cdot d \mathbb{Q}$. This negative term means Equation~\ref{equ:kl2} has a supreme value. Through setting the first-order functional derivative as zero $d \mathbb{P}-e^{f(u, v, y)} \cdot d \mathbb{Q}=0$, we can get the optimal mapping function $f^*(u, v, y)=\log \frac{d \mathbb{P}}{d \mathbb{Q}}$. Rewrite the Equation~\ref{equ:kl2} with $f^*$:
\begin{equation}
\begin{aligned}
    \mathbb{E}_{\mathbb{P}}\left[f^*(u, v, y)\right]-\mathbb{E}_{\mathbb{Q}}\left[e^{f^*(u, v, y)}\right]+1 & =\mathbb{E}_{\mathbb{P}}\left[\log \frac{d \mathbb{P}}{d \mathbb{Q}}\right]\\ & =D_{\mathrm{KL}}(\mathbb{P} \| \mathbb{Q})
\end{aligned}
\end{equation}

\subsection{Results based on Lemma~\ref{lem:kl}}
\begin{lemma}
    \begin{equation}
\begin{aligned}
\text{Weak-CMI}&(U ; V \mid Y)\\ & =D_{\mathrm{KL}}\left(P_{U, V} \| \mathbb{E}_{P_Y}\left[P_{U \mid Y} P_{V \mid Y}\right]\right) \\
& =\sup _f \mathbb{E}_{(u, v) \sim P_{U, V}}[f(u, v)]\\
& \quad\quad\quad -\mathbb{E}_{(u, v) \sim \mathbb{E}_{P_Y}\left[P_{U \mid Y} P_{V \mid Y}\right]}\left[e^{f(u, v)}\right]+1
\end{aligned}
\end{equation}
\label{lem:weak_cmi}
\end{lemma}
\emph{Proof.} Let $\mathbb{P}$ be the joint distribution $P_{U,V}$ and $\mathbb{Q}$ be expectation on the product of marginal distribution $\mathbb{E}_{P_Y}\left[P_{U \mid Y} P_{V \mid Y}\right]$ in Lemma~\ref{lem:kl}.

\begin{lemma}
    \begin{equation}
    \begin{aligned}
&\sup _f \mathbb{E}_{\left(u, v_1\right) \sim \mathbb{P},\left(u, v_{2: n}\right) \sim \mathbb{Q}^{\otimes(n-1)}}\left[\log \frac{e^{f\left(u, v_1\right)}}{\frac{1}{n} \sum_{j=1}^n e^{f\left(u, v_j\right)}}\right] \\ & \leq D_{\mathrm{KL}}(\mathbb{P} \| \mathbb{Q})
\end{aligned}
\end{equation}
\label{lem:kl_long}
\end{lemma}
\emph{Proof.} $\forall f$, we can draw:
\begin{equation}
\begin{aligned}
D_{\mathrm{KL}}& (\mathbb{P} \| \mathbb{Q})  =\mathbb{E}_{\left(u, v_{2: n}\right) \sim \mathbb{Q}^{\otimes(n-1)}}\left[D_{\mathrm{KL}}(\mathbb{P} \| \mathbb{Q})\right] \\
& \geq \mathbb{E}_{\left(u, v_{2: n}\right) \sim \mathbb{Q}^{\otimes(n-1)}}\left[\mathbb{E}_{\left(u, v_1\right) \sim \mathbb{P}}\left[\log \frac{e^{f\left(u, v_1\right)}}{\frac{1}{n} \sum_{j=1}^n e^{f\left(u, v_j\right)}}\right] \right.\\
    &\quad\quad \left. -\mathbb{E}_{\left(u, v_1\right) \sim \mathcal{Q}}\left[\frac{e^{f\left(u, v_1\right)}}{\frac{1}{n} \sum_{j=1}^n e^{f\left(u, v_j\right)}}\right]+1\right] \\
& =\mathbb{E}_{\left(u, v_{2: n}\right) \sim \mathbb{Q} \otimes(n-1)}\Bigg[\mathbb{E}_{\left(u, v_1\right) \sim \mathbb{P}}\left[\log \frac{e^{f\left(u, v_1\right)}}{\frac{1}{n} \sum_{j=1}^n e^{f\left(u, v_j\right)}}\right]\\ & \quad\quad\quad\quad\quad\quad\quad\quad\quad -1+1\Bigg] \\
& =\mathbb{E}_{\left(u, v_1\right) \sim \mathbb{P},\left(u, v_{2: n}\right) \sim \mathbb{Q} \otimes(n-1)}\left[\log \frac{e^{f\left(u, v_1\right)}}{\frac{1}{n} \sum_{j=1}^n e^{f\left(u, v_j\right)}}\right] .
\end{aligned}
\end{equation}
In detail, the first line always exists because $D_{\mathrm{KL}}(\mathbb{P} \| \mathbb{Q})$ is a constant. The second line comes from Lemma~\ref{lem:kl}. And because $\left(u, v_1\right)$ and $\left(u, v_{2: n}\right)$ can be interchangeable when they are all sampled from $\mathbb{Q}$, we can obtain the result in the third line. In conclusion, since the inequality works for $\forall f$, we can obtain Lemma~\ref{lem:kl_long}

\subsection{Results based on Lemma~\ref{lem:kl2}}
\begin{lemma}
    \begin{equation}
\begin{aligned}
\text{CMI}(U ; V \mid Y)&:=\mathbb{E}_{P_Y}\left[D_{\mathrm{KL}}\left(P_{U, V \mid Y} \| P_{U \mid Y} P_{V \mid Y}\right)\right] \\
& =D_{\mathrm{KL}}\left(P_{U, V, Y} \| P_Y P_{U \mid Y} P_{V \mid Y}\right) \\
& =\sup _f \mathbb{E}_{(u, v, y) \sim P_{U, V, Y}}[f(u, v, y)]\\ &\quad -\mathbb{E}_{(u, v, y) \sim P_Y P_{U \mid Y} P_{V \mid Y}}\left[e^{f(u, v, y)}\right]+1 \\
&
\end{aligned}
\end{equation}
\label{lem:cmi}
\end{lemma}

\subsection{Proving $\text { Weak-CMI }(U;V \mid Y) \leq \text{CMI}(U;V \mid Y)$}
\begin{proposition}
    $\text { Weak-CMI }(U;V \mid Y) \leq \text{CMI}(U;V \mid Y).$
\end{proposition}
\emph{Proof.} According to Lemma~\ref{lem:weak_cmi},
\begin{equation}
\label{pro:weak}
\begin{aligned}
\operatorname{Weak-CMI}&(U ; V \mid Y)\\  & =\sup _f \mathbb{E}_{(u, v) \sim P_{U, V}}[f(u, v)]\\ & \quad\quad -\mathbb{E}_{(u, v) \sim \mathbb{E}_{P_Y}\left[P_{U \mid Y} P_{V \mid Y}\right]}\left[e^{f(u, v)}\right]+1 \\
& =\sup _f \mathbb{E}_{(u, v, y) \sim P_{U, V, Y}}[f(u, v)]\\ & \quad\quad -\mathbb{E}_{(u, v, y) \sim P_Y P_{U \mid Y} P_{U \mid Y}}\left[e^{f(u, v)}\right]+1
\end{aligned}
\end{equation}
When the equality for $\operatorname{Weak-CMI}$ holds, we assume the function as $f_1^*(u, v)$. And let $f_2^*(u, v, y)=f_1^*(u, v)$ which means $\forall y \sim P_Y$, $f_2^*(u, v, y)$ will not change. Then, we can get:
\begin{equation}
\begin{aligned}
\operatorname{Weak-CMI}&(U ; V \mid Y) \\  &=\mathbb{E}_{(u, v, y) \sim P_{U, V, Y}}\left[f_1^*(u, v)\right] \\ & \quad\quad -\mathbb{E}_{(u, v, y) \sim P_Y P_{U \mid Y} P_{V \mid Y}}\left[e^{f_1^*(u, v)}\right]+1 \\
& =\mathbb{E}_{(u, v, y) \sim P_{U, V, Y}}\left[f_2^*(u, v, y)\right]\\ & \quad\quad -\mathbb{E}_{(u, v, y) \sim P_Y P_{U \mid Y} P_{U \mid Y}}\left[e^{f_2^*(u, v, y)}\right]+1
\end{aligned}
\label{equ:weak}
\end{equation}

Comparing Equation~\ref{equ:weak} with Lemma~\ref{lem:cmi}, we can conclude $\text { Weak-CMI }(U;V \mid Y) \leq \text{CMI}(U;V \mid Y)$.

\subsection{Showing the Equation~\ref{equ:con} is a lower bound of CMI}
\begin{proposition}
    We restate the Equation~\ref{equ:con} in the main text, and call it as the estimate of CMI (CMIE):
\begin{small}
\begin{equation}
\begin{aligned}
\text{CMIE} & :=\sup _f \underset{{y \sim P_Y}}{\mathbb{E}}\left[\mathbb{E}_{\left(u_i, v_i\right) \sim P_{U, V \mid y} \otimes n}\left[\log \frac{e^{f\left(u_i, v_i\right)}}{\frac{1}{n} \sum_{j=1}^n e^{f\left(u_i, v_j\right)}}\right]\right] \\
& \leq D_{\mathrm{KL}}\left(P_{U, V} \| \mathbb{E}_{P_Y}\left[P_{U \mid Y} P_{V \mid Y}\right]\right)\\ & =\text { Weak-CMI }(U ; V \mid Y) \leq \text{CMI}(U ; V \mid Y)
\end{aligned}
\end{equation}
\end{small}
\end{proposition}
\emph{Proof.} By defining $\mathbb{P}=P_{U,V}$ and $\mathbb{Q}=\mathbb{E}_{P_Y}\left[P_{U \mid Y} P_{V \mid Y}\right]$ we can obtain:
\begin{equation}
\begin{aligned}
\mathbb{E}_{\left(u, v_1\right) \sim \mathbb{P},\left(u, v_{2: n}\right) \sim \mathbb{Q} \otimes(n-1)}\left[\log \frac{e^{f\left(u, v_1\right)}}{\frac{1}{n} \sum_{j=1}^n e^{f\left(u, v_j\right)}}\right]= \\
\mathbb{E}_{y \sim P_Y}\left[\mathbb{E}_{\left(u_i, v_i\right) \sim P_{U, V \mid y} \otimes_n}\left[\log \frac{e^{f\left(u_i, v_i\right)}}{\frac{1}{n} \sum_{j=1}^n e^{f\left(u_i, v_j\right)}}\right]\right]
\end{aligned}
\end{equation}
Combined with Lemma~\ref{lem:kl_long}, we can deduce:
\begin{equation}
\begin{aligned}
\sup _f \mathbb{E}_{y \sim P_Y}\left[\mathbb{E}_{\left(u_i, v_i\right) \sim P_{U, V \mid y}{ }^{\otimes n}}\left[\log \frac{e^{f\left(u_i, v_i\right)}}{\frac{1}{n} \sum_{j=1}^n e^{f\left(u_i, v_j\right)}}\right]\right] \leq \\
D_{\mathrm{KL}}\left(P_{U, V} \| \mathbb{E}_{P_Y}\left[P_{U \mid Y} P_{V \mid Y}\right]\right)
\end{aligned}
\end{equation}
Through Proposition~\ref{pro:weak} that $\text { Weak-CMI }(U;V \mid Y) \leq \text{CMI}(U;V \mid Y)$, we can draw the conclusion that CMIE is a lower bound of CMI. 

\section{Online Clustering\label{app:online}}
To avoid the trivial solution, we add the constraint that the prototype assignments must be equally partitioned following:
\begin{equation}
\mathcal{Q}=\left\{{Q} \in \mathbb{R}_{+}^{K \times B} \mid {Q} \mathds{1}_B=\frac{1}{K} \mathds{1}_K, {Q}^{\top} \mathds{1}_K=\frac{1}{B} \mathds{1}_B\right\},
\label{equ:contraint}
\end{equation}
where the matrix $Q=[q_{u_1}, q_{u_2}, \cdots, q_{u_B}]$ will be optimized that belong to \emph{transportation polytope}, $\mathds{1}_K, \mathds{1}_B$ denotes the vectors of all ones containing K, B dimension. Then, the objective function of Equation~\ref{equ:clu} can be reformulated as
\begin{equation}
\min _{p, q} \mathcal{L}_{\text {clu }}=\min _{Q \in \mathcal{Q}}\langle Q,-\log P\rangle_{\mathrm{F}}-\log B,
\label{equ:trans}
\end{equation}
where $P=[\frac{1}{B}p_{u_1},\cdots,\frac{1}{B}p_{u_B}]$ is calculated by Equation~\ref{equ:pq} and $\langle \cdot \rangle_{\mathrm{F}}$ is the Frobenius dot-product. The loss function in Equation~\ref{equ:trans} is an \emph{optimal transport problem} that can be efficiently addressed by iterative \emph{Sinkhorn-Knopp algorithm}~\cite{cuturi2013sinkhorn}:
\begin{equation}
{Q}^*=\operatorname{Diag}({t}) \exp \left(P^{\lambda}\right) \operatorname{Diag}({r}),
\end{equation}
where $t$ and $r$ are renormalization vectors which are calculated by the iterative Sinkhorn-Knopp algorithm, and the hyper-parameter $\lambda$ is employed to balance the convergence speed and the proximity to the original transport problem.

\section{Experimental Details}
\subsection{Baselines\label{app:baselines}}
We consider empirical risk minimization (ERM), one OOD algorithm IRM and one graph-specific OOD algorithm EERM as supervised baselines. And we include 9 self-supervised methods as unsupervised baselines:
\begin{itemize}
    \item Invariant Risk Minimization (IRM~\cite{irm}) is an algorithm that seeks to learn data representations that are robust and generalize well across different environments by penalizing feature distributions that have different optimal linear classifiers for each environment
    \item EERM~\cite{eerm}  generates multiple graphs by environment generators and minimizes the mean and variance of risks from multiple environments to capture invariant features.
    \item Graph Autoencoder (GAE~\cite{gae}) is an encoder-decoder structure model. Given node attributes and structures, the encoder will compress node attributes into low-dimension latent space, and the decoder (dot-product) hopes to reconstruct existing links with compact node features.
    \item Variational Graph Autoencoder (VGAE~\cite{gae}) is similar to GAE but the node features are re-sampled from a normal distribution through a re-parameterization trick.
    \item GraphMAE~\cite{gmae} is a masked autoencoder. Different to GAE and VGAE, it will mask partial input node attributes firstly and then the encoder will compress the masked graph into latent space, finally a decoder aims to reconstruct the masked attributes. 
    \item Deep Graph Infomax (DGI~\cite{dgi}) is a node-graph contrastive method which contrasts the node representations and graph representation. First, it will apply the corrupt function to obtain a negative graph and two graphs will be fed into a shared GNN model to generate node embeddings. And a readout function will be applied on the original node embeddings to obtain graph-level representation. Corrupted embeddings and readout graph representation are considered as positive pairs, original node representations and readout graph representation are considered as positive pairs. 
    \item MVGRL~\cite{mvgrl} is similar to DGI but utilizes the information of multi-views. Firstly, it will use edge diffusion function to generate an augmented graph. And asymmetric encoders will be applied on the original graph and diffusion graph to acquire node embeddings. Next, a readout function is employed to derive graph-level representations. Original node representations and augmented graph-level representation are regarded positive pairs. Additionally, the augmented node representations and original graph-level representation are also considered as positive pairs. The negative pairs are constructed following \cite{dgi}.
    \item RoSA~\cite{rosa} is a robust self-aligned graph contrastive framework which does not require the explicit alignment of nodes in the positive pairs so that allows more flexible graph augmentation. It proposes the graph earth move distance (g-EMD) to calculate the distance between unaligned views to achieve self-alignment. Furthermore, it will use adversarial training to realize robust alignment.
    \item GRACE~\cite{grace} is node-node graph contrastive learning method. It designs two augmentation functions (\ie, removing edges and masking node features) to generate two augmented views. Then a shared graph model will be applied on augmented views to generate node embedding matrices. The node representations augmented from the same original node are regarded as positive pairs, otherwise are negative pairs. Lastly, pairwise loss (\eg, InfoNCE~\cite{cpc}) will be applied on these node matrices.
    \item BGRL~\cite{bgrl} is similar to GRACE but without negative samples which is motivated by BYOL~\cite{byol}.
    \item COSTA~\cite{costa} proposes feature augmentation to decrease the bias introduced by graph augmentation.
    \item SwAV~\cite{swav} is an unsupervised online clustering method which incorporates prototypes for clustering and employs swapped prediction for model training. It is originally designed for the computer vision domain, we adopt it into graph domain.
\end{itemize}

\subsection{Datasets\label{app:datasets}}
For GOOD-Cora, GOOD-Twitch, GOOD-CBAS and GOOD-WebKB datasets, they are all adopted from GOOD\cite{good} which is a comprehensive Graph OOD benchmark. These datasets contain both concept shift and covariate shift splits, for more details of splitting, please refer to Appendix A in \cite{good}. 

GOOD-Cora is a citation network that is derived from the full Cora dataset~\cite{bojchevskideep}. In the network, each node represents a scientific publication, and edges between nodes denote citation links. The task is to predict publication types (70-classification) of testing nodes. The data splits are generated based on two domain selections (\ie, word, degree). 
The word diversity selection is based on the count of selected words within a publication and is independent of the publication's label. On the other hand, the node degree selection ensures that the popularity of a paper does not influence its assigned class.

GOOD-Twitch is a gamer network dataset. In this dataset, each node represents a gamer, and the node features correspond to the games played by each gamer. The edges between nodes represent friendship connections among gamers. The binary classification task associated with this dataset involves predicting whether a user streams mature content or not. The data splits for GOOD-Twitch are based on the user language, ensuring that the prediction target is not biased by the specific language used by a user.

GOOD-CBAS is a synthetic dataset that is modified from the BA-Shapes dataset~\cite{ying2019gnnexplainer}. It involves a graph where 80 house-like motifs are attached to a base graph following the Barabási–Albert model, resulting in a graph with 300 nodes. The task associated with this dataset is to predict the role of each node within the graph. The roles can be classified into four classes, which include identifying whether a node belongs to the top, middle, or bottom of a house-like motif, or if it belongs to the base graph itself.
In contrast to using constant node features, the GOOD-CBAS dataset introduces colored features. This modification poses challenges for out-of-distribution (OOD) algorithms, as they need to handle differences in node colors within covariate splits and consider the correlations between node color and node labels within concept splits.

GOODWebKB is a network dataset that focuses on university webpages. Each node in the network represents a webpage, and the node features are derived from the words that appear on the webpage. The edges between nodes represent hyperlinks between webpages. The task associated with this dataset is a 5-class prediction task, where the goal is to predict the class of each webpage. The data splits for GOOD-WebKB are based on the domain of the university, ensuring that the classification of webpages is based on their word contents and link connections rather than any specific university features.

Amazon-Photo is a co-purchasing network that is widely used for evaluating the design of GNN models. In this network, each node corresponds to a specific product, and the presence of an edge between two products indicates that they are frequently purchased together by customers. In the original dataset, it is observed that the node features exhibit a significant correlation with the corresponding node labels. In order to evaluate the model's ability to generalize to out-of-distribution scenarios, it is necessary to introduce distribution shifts into the training and testing data. To achieve this, we adopt the strategies employed in the EERM~\cite{eerm}. Specifically, we leverage the available node features $X_1$ to create node labels $Y$ and spurious environment-sensitive features $X_2$. To elaborate, a randomly initialized GNN takes $X_1$ and the adjacency matrix as inputs and employs an argmax operation in the output layer to obtain one-hot vectors as node labels. Additionally, we employ another randomly initialized GNN that takes the concatenation of $Y$ and an environment id as input to generate spurious node features $X_2$. By combining these two sets of features, we obtain the input node features, $X = [X_1, X_2]$, which are used for both training and evaluation. This process is repeated to create ten graphs with distinct environment id's. Such a shift between different graphs can be considered as a concept shift~\cite{ood-survey}. Finally, one graph is allocated for training, another for validation, and the remaining graphs are used for evaluating the OOD generalization of the trained model.

Elliptic is a financial network that records the payment flows among transactions as time goes by. It consists of 49 graph snapshots which are collected at different times. Each graph snapshot is a network of Bitcoin transactions where each node represents one transaction and each edge denotes a payment flow. Partial nodes (approximately 20\%) are labeled as licit or illicit transactions and we hope to identify illicit transactions in the future. For data preprocessing, we adopt the same strategies in EERM~\cite{eerm}: removing extremely imbalanced snapshots and using the 7th-11th/12th-17th/17th-49th snapshots for training/validation/testing data. And 33 testing graph snapshots will be split into 9 test sets according to chronological order. In Figure~\ref{fig:rate}, we depict the label rate and positive label rate for training/validation/testing sets. It is evident that the varying positive label rates across different data sets are apparent. Indeed, the model needs to deal with the label distribution shifts from training to testing data.
\begin{figure}
    \centering
    \includegraphics[scale=0.4]{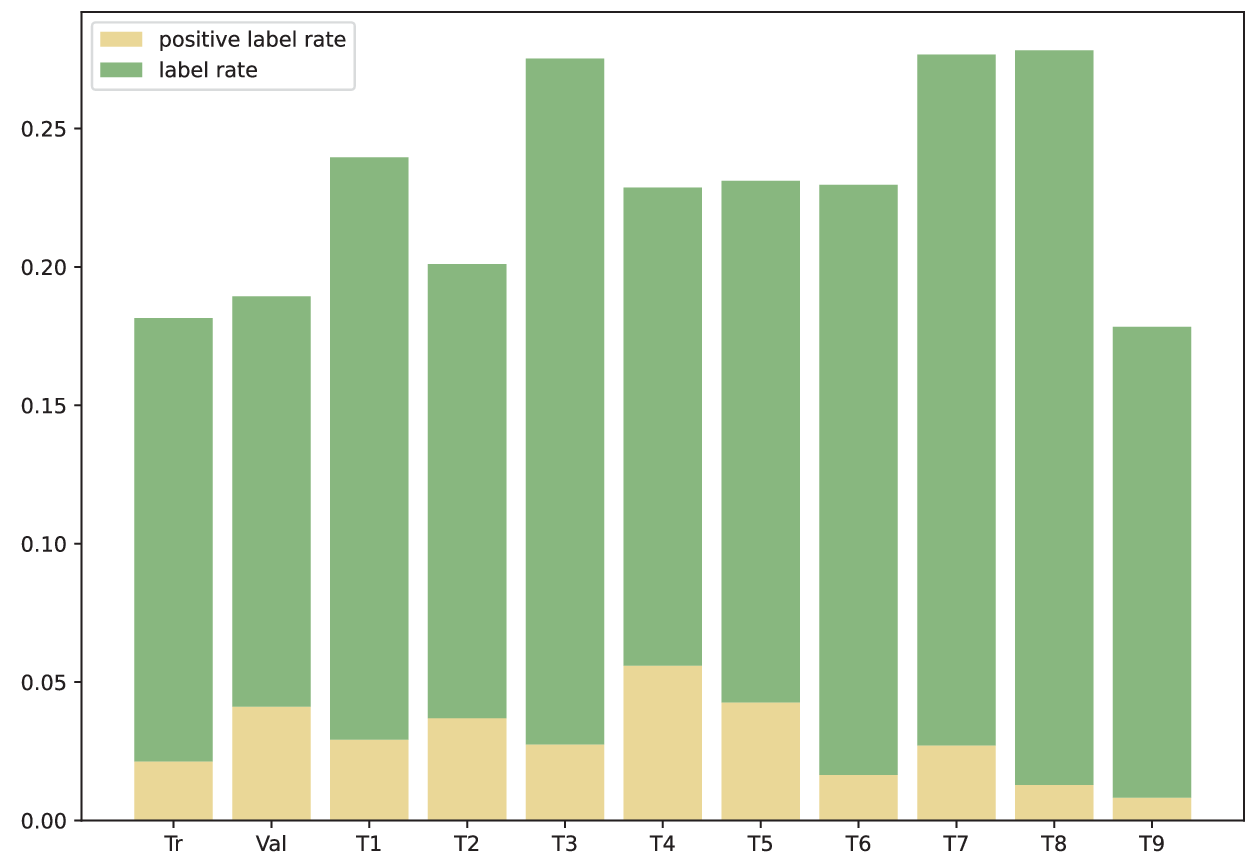}
    \caption{The label rates and positive label rates of Elliptic dataset. `Tr' means training data, `Val' denotes validation data and others represent testing data. In the different splits, the label distributions are disparate.}
    \label{fig:rate}
\end{figure}
\subsection{Hyper-parameters~\label{app:hyper}}
For GOOD datasets, we adopt GraphSAINT~\cite{} as subsampling technique, while utilizing a 3-layer GCN~\cite{gcn} with 300 hidden units as the backbone following~\cite{good}. For the supervised baselines (\ie, ERM, IRM, EERM), we use the identical hyper-parameters specified in \cite{good}. For other unsupervised baselines, we conduct a grid search to find the best performance. Specifically, max training epoch ranges in \{50, 100, 200, 500, 600\} and learning rate ranges in \{1e-1,1e-2,1e-3,1e-4,1e-5\}, augmentation ratio range in $[0.1, 0.6]$. Regarding GraphMAE, the masking ratio ranges in \{0.25,0.5,0.75\}, and we use a one-layer GCN as the decoder. For MARIO, the specific hyper-parameters are listed in Table~\ref{tab:hyper}. For the adversarial augmentation, we set ascent steps $M$ as 3 and the ascent step size $\epsilon$ as 1e-3. For detailed hyper-parameters of all algorithms, please refer to \hyperlink{https://github.com/ZhuYun97/MARIO}{https://github.com/ZhuYun97/MARIO}.

For Amazon-Photo, we utilize 2-layer GCN with 128 hidden units as encoder, and we set $\tau, p_{f,1}, p_{f,2}, p_{e,1}, p_{e,2}, \gamma, |C|$ as 0.2, 0.2, 0.3, 0.2, 0.3, 0.1, 100 respectively and learning rate as 1e-4. Other hyper-parameters remain the same in EERM~\cite{eerm}.
Regarding Elliptic datasets, we employ 5-layer GraphSAGE~\cite{sage} with 32 hidden units as encoder following EERM~\cite{eerm}. And we set $\tau, p_{f,1}, p_{f,2}, p_{e,1}, p_{e,2}, \gamma, |C|$ as 0.8, 0.2, 0.3, 0.2, 0.3, 0.5, 120 respectively. The remaining hyper-parameters are consistent with EERM~\cite{eerm}.
\begin{table*}[htp]
\caption{Hyperparameters specifications for MARIO}
    \label{tab:hyper}
    \centering
    \begin{tabular}{c|cc|cc|cc|cc|cc}
    \toprule
              & \multicolumn{4}{c|}{GOOD-Cora}                                     & \multicolumn{2}{c|}{GOOD-CBAS}  & \multicolumn{2}{c|}{GOOD-Twitch} & \multicolumn{2}{c}{GOOD-WebKB} \\
              & \multicolumn{2}{c}{word}         & \multicolumn{2}{c|}{degree}     & \multicolumn{2}{c|}{color}      & \multicolumn{2}{c|}{language}    & \multicolumn{2}{c}{university}  \\
              & concept      & covariate         & concept      & covariate       & concept      & covariate       & concept      & covariate        & concept      & covariate       \\     \midrule                  
Model         & GCN          & GCN               & GCN          & GCN             & GCN          & GCN             & GCN          & GCN              & GCN          & GCN             \\
\# Layers     & 3            & 3                 & 3            & 3               & 3            & 3               & 3            & 3                & 3            & 3                \\
\# Hidden size & 300         & 300               & 300          & 300             & 300          & 300             & 300          & 300              & 300          & 300              \\
Epochs        & 100          & 150               & 100          & 100             & 200          & 500             & 600          & 200              & 100          & 500               \\
Learning rate & 1e-3         & 1e-3              & 1e-3         & 1e-3            & 1e-1         & 1e-2            & 1e-1         & 1e-1             & 1e-2         & 1e-2               \\ \midrule
$\tau$        & 0.2          & 0.2               & 0.2          & 0.2             & 0.2          & 0.2             & 0.2          & 0.2              & 0.5          & 0.5                 \\
$p_{f,1}$     & 0.3          & 0.3               & 0.0          & 0.3             & 0.2          & 0.2             & 0.2          & 0.2              & 0.5          & 0.2                \\
$p_{f,2}$     & 0.4          & 0.3               & 0.0          & 0.3             & 0.3          & 0.3             & 0.3          & 0.3              & 0.5          & 0.3                   \\
$p_{e,1}$     & 0.4          & 0.4               & 0.6          & 0.6             & 0.2          & 0.2             & 0.2          & 0.2              & 0.5          & 0.2                \\
$p_{e,2}$     & 0.5          & 0.4               & 0.6          & 0.6             & 0.3          & 0.3             & 0.3          & 0.3              & 0.5          & 0.3                \\
$|C|$         & 100          & 100               & 150          & 150             & 100          & 100             & 100          & 100              & 100          & 100               \\
$\gamma$      & 0.2          & 0.5               & 0.8          & 0.8             & 0.1          & 0.1             & 0.2          & 0.2              & 0.1          & 0.2                  \\ 
Pro. LR       & 1e-5         & 1e-5              & 1e-5         & 1e-5            & 1e-4         & 1e-3            & 1e-5         & 1e-3             & 1e-3         & 1e-3                \\\midrule
Epochs for LC & 500          & 200               & 100          & 100             & 100          & 2000            & 100          & 100              & 50           & 50                 \\
LR for LC     & 1e-4         & 1e-3              & 1e-3         & 1e-3            & 1e-2         & 1e-3            & 1e-1         & 1e-1             & 1e-1         & 1e-1                 \\
\bottomrule
    \end{tabular} 
    
\end{table*}

\subsection{Evaluation metrics}
In order to evaluate the pre-trained models, we adopt the linear evaluation protocol which is commonly used in self-supervised methods~\cite{simclr,moco,byol}. That is, we will train a linear classifier (\ie, one-layer MLP) on top of (frozen) representations learned by self-supervised methods. The training epochs (Epochs for LC in Table~\ref{tab:hyper}) and learning rate (LR for LC in Table~\ref{tab:hyper}) of the linear classifier are obtained by grid search.

\subsection{Computer infrastructures specifications}
For hardware, all experiments are conducted on a computer server with eight GeForce RTX 3090 GPUs with 24GB memory and 64 AMD EPYC 7302 CPUs. And our models are implemented by Pytorch Geometric 2.0.4~\cite{fey2019fast} and Pytorch 1.11.0~\cite{torch}. All datasets used in our
work are available on \hyperlink{https://github.com/divelab/GOOD}{https://github.com/divelab/GOOD} and \hyperlink{https://github.com/qitianwu/GraphOOD-EERM}{https://github.com/qitianwu/GraphOOD-EERM}.

\section{Additional Experiments}
\subsection{Graph classification~\label{app:graph_classification}}

\subsubsection{Datasets}
The PROTEINS dataset comprises protein data. During training, we use graphs ranging from 4 to 25 nodes, while during testing, we evaluate on graphs spanning from 6 to 620 nodes. 
The D\&D dataset is also protein-based and involves two distinct splitting methods, namely $\text{D\&D}_{200}$ and $\text{D\&D}_{300}$. 
For the $\text{D\&D}_{200}$ split, training is conducted on graphs containing 30 to 200 nodes, and testing is performed on graphs consisting of 201 to 5,748 nodes. As for the $\text{D\&D}_{300}$ split, training is carried out on 500 graphs ranging from 30 to 300 nodes, while testing is conducted on other graphs comprising 30 to 5,748 nodes.
\subsubsection{Experimental setups \& Baselines}
\textbf{Experimental setups.} For all graph classification datasets, we utilize 2-layer GCN containing 64 hidden units as graph encoder, and we choose global max pooling as the readout function. For other hyper-parameters, we set $\text{lr}, \tau, p_{f,1}, p_{f,2}, p_{e,1}, p_{e,2}, \gamma, |C|$ as 0.01, 0.5, 0.2, 0.3, 0.2, 0.3, 0.5, 40 respectively for self-supervised methods. For ERM, the learning rate ranges in \{1e-2, 1e-3, 5e-3, 1e-4\} and the number of training epochs is selected in \{50, 100, 200\}.
For evaluating pre-trained models, we use an off-the-shelf $\ell_2$-regularized LogisticRegression classifier from Scikit-Learn ~\cite{pedregosa2011scikit} using the 'liblinear' solver with a small hyperparameter search over the regularization strength to be between $\left\{2^{-10}, 2^{-9}, \ldots 2^9, 2^{10}\right\}$.

\noindent\textbf{Baselines.} GraphCL~\cite{graphcl} investigated the impact of different graph augmentations (\ie, node dropping, edge perturbation, attribute masking and subgraph sampling) on graph classification datasets. The framework is similar to  SimCLR~\cite{simclr} but specified for graph domain.

\subsubsection{Results}
From Table~\ref{tab:graph_class}, we can find GraphCL~\cite{graphcl} can achieve comparable performance with ERM even without labels. And our recipe can boost the OOD generalization ability of unsupervised methods and even surpasses ERM which means our recipe is also effective for graph classification task.
\begin{table}[htp]
\caption{Results of different methods on OOD graph classification tasks. We report the mean of Accuracy with standard deviation after 10 runs.}
    \label{tab:graph_class}
    \centering
    \begin{tabular}{c|ccc}
    \toprule
                    & $\text{PROTEIN}_{25}$ & $\text{D\&D}_{200}$ & $\text{D\&D}_{300}$\\ \midrule
\#Train/Test Graphs &  500/613              & 462/716         & 400/678        \\
\#Nodes Train       & 4-25                   & 30-200          & 30-300         \\
\#Nodes Test        & 6-620                  & 201-5748        & 30-5748        \\ \midrule
    ERM             & 77.24±0.95             & 44.25±5.16      & 67.91±1.60       \\
    GraphCL         & 76.92±0.91             & 48.12±6.43      & 67.82±1.29     \\
    GraphCL(+MARIO) & 78.08±0.97             & 51.62±5.47      & 69.13±1.23     \\ \bottomrule
    \end{tabular}
    
\end{table}

\subsection{Integrated with other methods~\label{app:other_methods}}
Our recipe is not only model-agnostic but also an add-on training scheme that can be adopted on most graph contrastive learning (GCL) methods. In Table~\ref{tab:other_methods}, we use our recipe to guide various GCL methods (GRACE, COSTA). MARIO can further boost these methods on both ID and OOD test performance.
\begin{table}[htp]
\caption{Results of various methods integrated with MARIO. We report the mean and standard deviation of Accuracy after 10 runs.}
    \label{tab:other_methods}
\scalebox{0.9}{
    \centering
    \begin{tabular}{c|cc|cc}
    \toprule
        \multirow{3}{*}{concept shift}  & \multicolumn{2}{c|}{GOOD-WebKB}  &  \multicolumn{2}{c}{GOOD-CBAS} \\
                 & \multicolumn{2}{c|}{university}  & \multicolumn{2}{c}{color}      \\ 
                 &        ID  & OOD                &    ID        & OOD             \\ \midrule
        GRACE    & 64.00±3.43 & 34.86±3.43         &  92.00±1.39  & 88.64±0.67       \\
GRACE (+MARIO)   & 65.67±2.81 & 37.15±2.37         &  94.36±1.21  & 91.28±1.10      \\ \midrule
        COSTA    & 61.66±2.58 & 32.39±2.13         &  93.50±2.62  & 89.29±3.11      \\
COSTA(+MARIO)    & 62.33±2.60 & 35.32±3.46         &  98.00±1.31  & 94.36±1.51      \\ \bottomrule
    \end{tabular} }
    
\end{table}
\subsection{Metric scores curves\label{app:curves}}
In the main text, we draw the metric scores curves of GOOD-CBAS datasets, here we draw more metric scores curves. We can draw the same conclusions described in the main text from Figure~\ref{fig:curve_cora1},\ref{fig:curve_cora2},\ref{fig:curve_twitch},\ref{fig:curve_webkb}.
\begin{figure}
    \centering
    \includegraphics[scale=0.45]{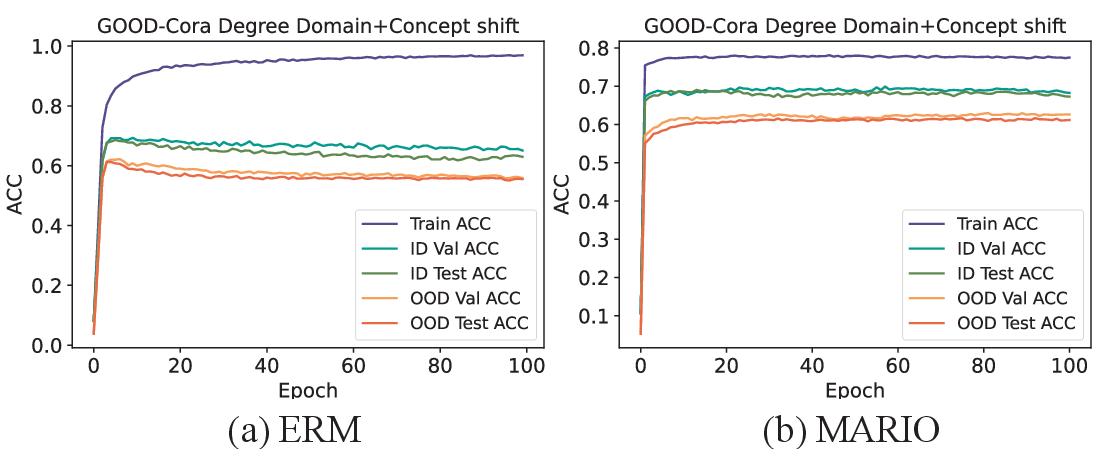}
    \caption{Metric score curves for ERM and MARIO on GOOD-Cora word domain with concept shift.}
    \label{fig:curve_cora1}
\end{figure}
\begin{figure}
    \centering
    \includegraphics[scale=0.45]{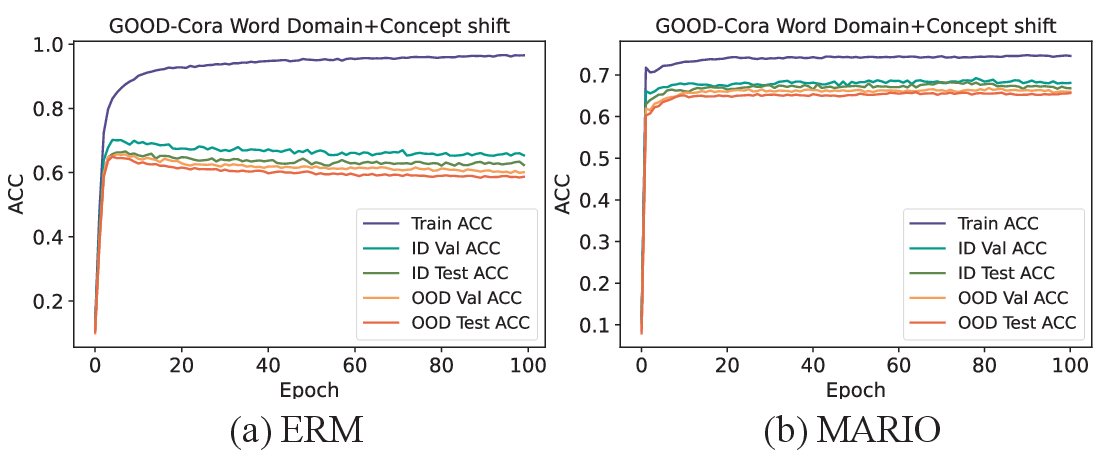}
    \caption{Metric score curves for ERM and MARIO on GOOD-Cora degree domain with concept shift.}
    \label{fig:curve_cora2}
\end{figure}
\begin{figure}
    \centering
    \includegraphics[scale=0.45]{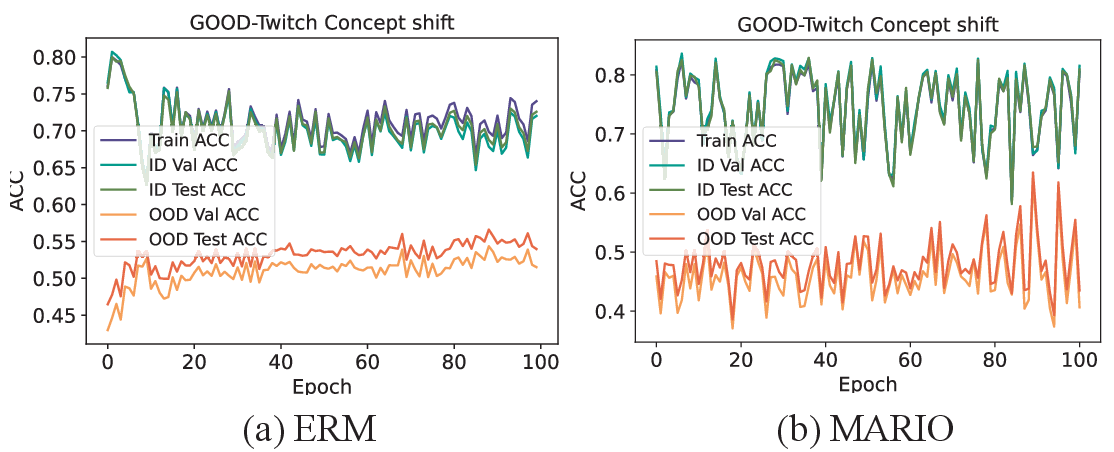}
    \caption{Metric score curves for ERM and MARIO on GOOD-Twitch language domain with concept shift.}
    \label{fig:curve_twitch}
\end{figure}
\begin{figure}
    \centering
    \includegraphics[scale=0.5]{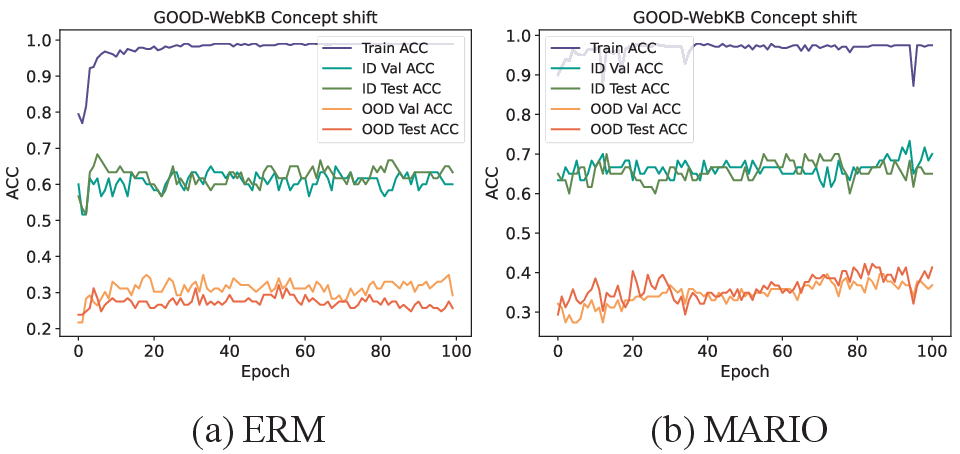}
    \caption{Metric score curves for ERM and MARIO on GOOD-WebKB university domain with concept shift.}
    \label{fig:curve_webkb}
\end{figure}

\subsection{Feature Visualization\label{app:feature}}
In the main text, we visualize all node (including ID and OOD nodes) embeddings in the same figure, we can find our method can map ID and OOD nodes in a similar feature space and the margin of each cluster is larger which can prove the superiority of our method dealing with OOD data. In this subsection, we separately draw ID and OOD nodes in different figures. In Figure~\ref{fig:tsne_id_ood}, the first the depict the ID node embeddings and the second line draws the OOD node embeddings. We can also observe that MARIO performs better clustering compared to other methods.
\begin{figure}[h]
    \centering
    \includegraphics[scale=0.35]{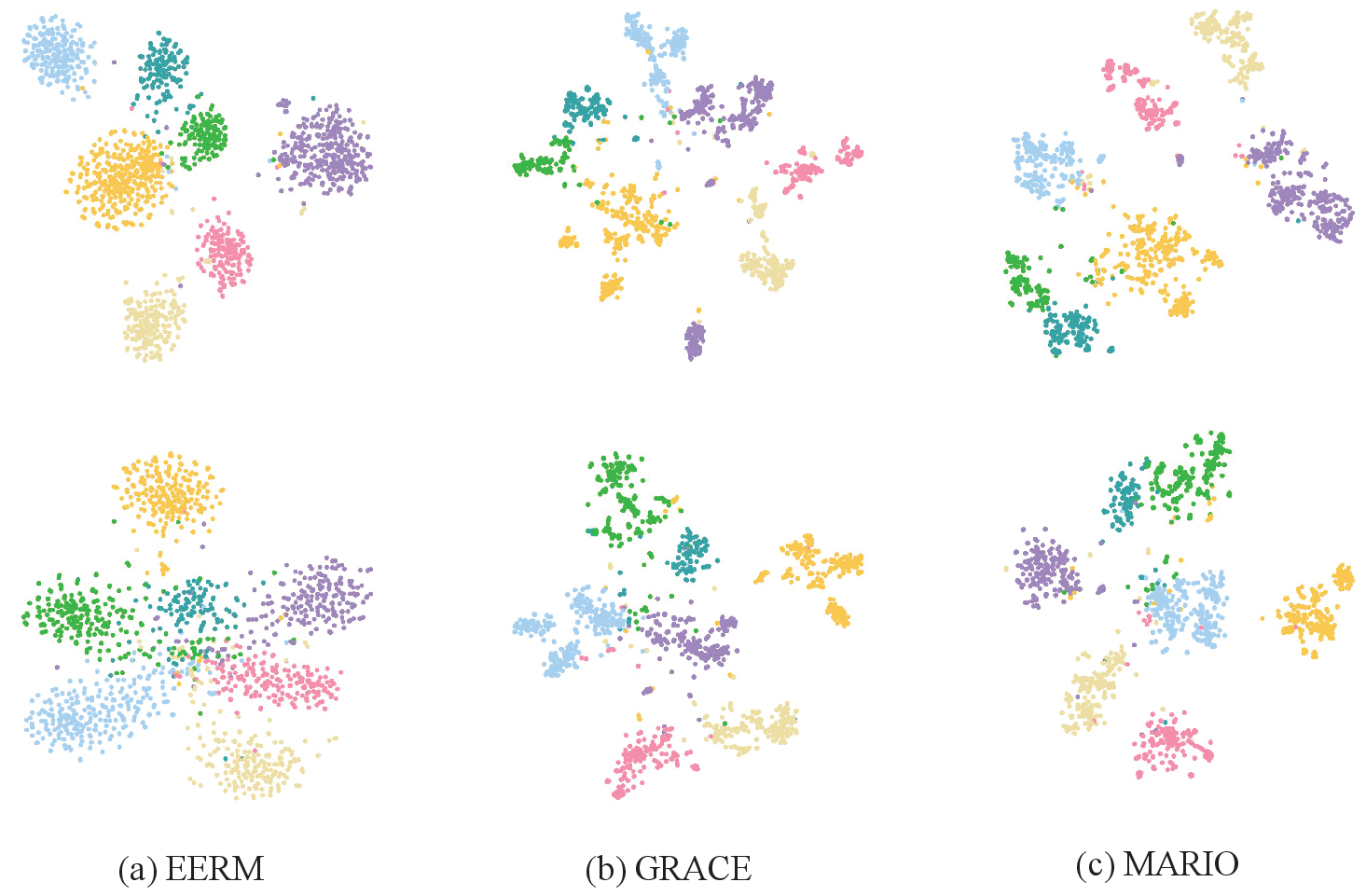}
    \caption{t-SNE visualization of node embeddings on GOOD-Cora dataset\protect\footnotemark, (a) depicts node embeddings from trained EERM, (b) shows embeddings from trained GRACE model, (c) is the result of trained MARIO. The margins of each cluster learned from MARIO are much wider than others.}
    \label{fig:tsne_id_ood}
\end{figure}
\footnotetext{Different from the main text, we select seven informative classes. And all models are trained under concept shift in degree domain.}

}
\end{document}